\documentclass[acmtog, nonacm]{acmart}

\usepackage{xcolor}
\usepackage{caption}
\usepackage{subcaption}
\usepackage{enumitem}
\usepackage{multirow}
\usepackage{makecell}
\usepackage{mathtools}
\usepackage{soul}
\usepackage{siunitx}
\AtBeginDocument{%
  \providecommand\BibTeX{{%
    \normalfont B\kern-0.5em{\scshape i\kern-0.25em b}\kern-0.8em\TeX}}}

\usepackage{xfrac}

\newcommand{\norm}[1]{\left\lVert#1\right\rVert}

\usepackage[makeroom]{cancel}

\newcommand{\density}{\tau}
\newcommand{\colordiffuse}{\mathbf{c_d}}

\newcommand\commentout[1]{}

\usepackage{colortbl}
\definecolor{yellow}{rgb}{1,1, 0.7}
\definecolor{gold}{rgb}{.9, 0.63, 0.}
\definecolor{lightyellow}{rgb}{1,1, 0.8}
\definecolor{orange}{rgb}{1, 0.85, 0.7}
\definecolor{tablered}{rgb}{1, 0.7, 0.7}
\usepackage[capitalize]{cleveref}
\crefname{section}{Sec.}{Secs.}
\Crefname{section}{Section}{Sections}
\Crefname{table}{Table}{Tables}
\crefname{table}{Tab.}{Tabs.}

\newcommand{\newcontract}{\operatorname{contract_\pi}}

\usepackage{expl3}
\ExplSyntaxOn
\newcommand\latinabbrev[1]{
  \peek_meaning:NTF . {
    #1\@}%
  { \peek_catcode:NTF a {
      #1.\@ }%
    {#1.\@}}}
\ExplSyntaxOff

\def\ie{\latinabbrev{i.e}}

\begin{document}

\title[MERF: Memory-Efficient Radiance Fields for Real-time View Synthesis in Unbounded Scenes]{MERF: Memory-Efficient Radiance Fields for \\ Real-time View Synthesis in Unbounded Scenes}

\author{Christian Reiser}
\authornote{Work done while interning at Google.}
\affiliation{
\institution{University of Tübingen, Tübingen AI Center}
\country{Germany}
}
\affiliation{
\institution{Google Research}
\country{United Kingdom}
}

\author{Richard Szeliski}
\affiliation{
\institution{Google Research}
\country{United States of America}
}

\author{Dor Verbin}
\affiliation{
\institution{Google Research}
\country{United States of America}
}

\author{Pratul P. Srinivasan}
\affiliation{
\institution{Google Research}
\country{United States of America}
}

\author{Ben Mildenhall}
\affiliation{
\institution{Google Research}
\country{United States of America}
}

\author{Andreas Geiger}
\affiliation{
\institution{University of Tübingen, Tübingen AI Center}
\country{Germany}
}

\author{Jonathan T. Barron}
\affiliation{
\institution{Google Research}
\country{United States of America}
}

\author{Peter Hedman}
\affiliation{
\institution{Google Research}
\country{United Kingdom}
}


\begin{abstract}
Neural radiance fields enable state-of-the-art photorealistic view synthesis. However, existing radiance field representations are either too compute-intensive for real-time rendering or require too much memory to scale to large scenes. We present a Memory-Efficient Radiance Field (MERF) representation that achieves real-time rendering of large-scale scenes in a browser. MERF reduces the memory consumption of prior sparse volumetric radiance fields using a combination of a sparse feature grid and high-resolution 2D feature planes. To support large-scale unbounded scenes, we introduce a novel contraction function that maps scene coordinates into a bounded volume while still allowing for efficient ray-box intersection. 
We design a lossless procedure for baking the parameterization used during training into a model that achieves real-time rendering while still preserving the photorealistic view synthesis quality of a volumetric radiance field.
\end{abstract}

\makeatletter
\let\@authorsaddresses\@empty
\makeatother

\begin{teaserfigure}
\begin{center}
  {\huge Interactive web demo at  \textcolor{blue}{\texttt{\href{https://merf42.github.io}{https://merf42.github.io}}}}\\
  \vspace{0.4cm}
  \includegraphics[width=\linewidth]{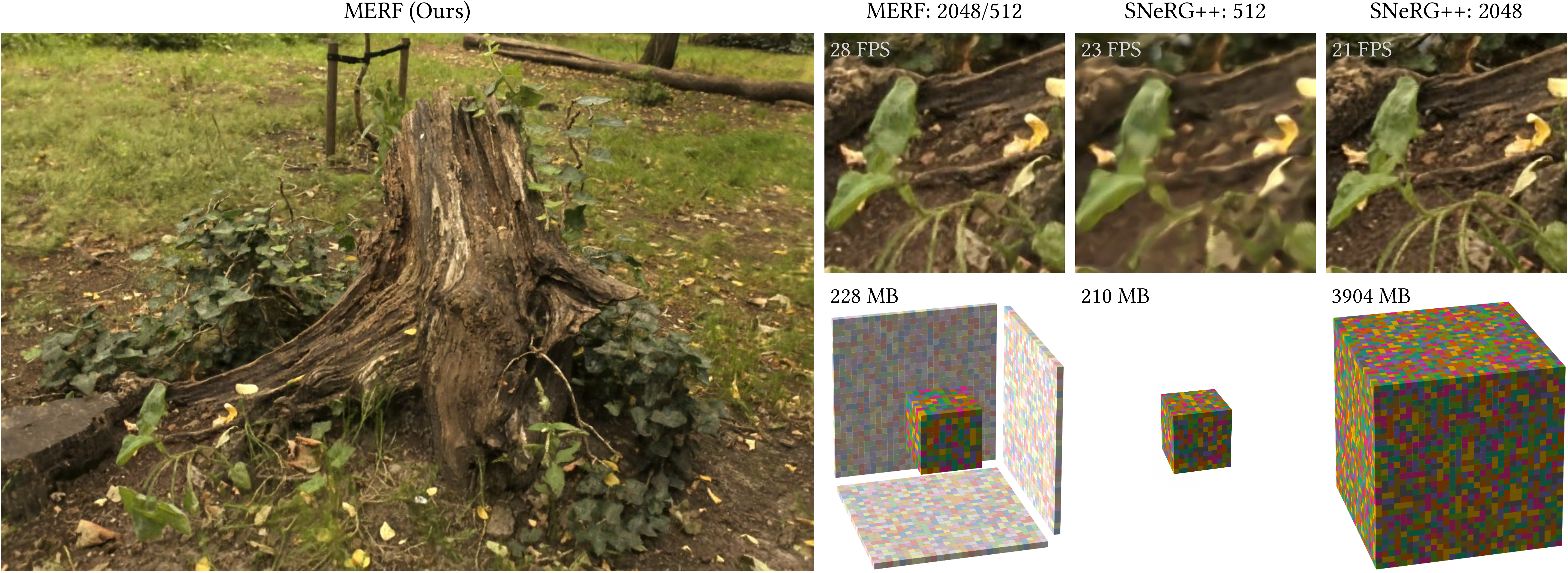}
  \caption{
  Our method, \emph{MERF}, allows for real-time view synthesis in the browser (framerates shown here are measured on a Macbook Pro M1). 
  We design a volumetric scene representation that is first optimized to maximize rendering quality and then losslessly baked into a more efficient format for fast rendering. 
  As depicted above, this representation uses a combination of a low-resolution 3D grid and a set of higher-resolution 2D planes (spatial extent of each visualization signifies memory consumption).
  Compared to prior real-time view synthesis methods such as SNeRG++ (our improved version of \citet{hedman2021snerg}), MERF achieves better image quality for the same amount of compression, matching the output of a SNeRG++ model that is over 17$\times$ larger.
  }
  \label{fig:teaser}
\end{center}
\end{teaserfigure}

\maketitle

\section{Introduction}
Neural volumetric scene representations such as Neural Radiance Fields (NeRF)~\cite{mildenhall2020nerf} enable photorealistic novel view synthesis of scenes with complex geometry and appearance, but the compute required to query such neural representations during volumetric raymarching prohibits real-time rendering. Subsequent works have proposed discretized volumetric representations that can substantially increase rendering performance~\cite{garbin2021fastnerf,yu2021plenoctrees,hedman2021snerg}, but these approaches do not yet enable practical real-time rendering for \emph{large-scale} scenes. 

These representations struggle to scale to larger scenes primarily due to graphics hardware constraints.
Volumetric data is necessarily larger than a 2D surface representation and occupies more space in memory.
Similarly, while a camera ray intersects a hard surface at most once, rendering a ray through a volume may require many samples.
With state of the art neural or hybrid representations, each of these sample queries is very expensive to evaluate, either in terms of compute or memory bandwidth.
As a result, methods that work for scenes with limited extent (single objects in space or forward-facing scenes) typically do not scale up to larger unbounded scenes.

All neural or hybrid volumetric methods must address two fundamental trade-offs that arise from these constraints: 
\begin{itemize}[noitemsep,topsep=0pt,parsep=0pt,partopsep=0pt,leftmargin=*]
\item 
\emph{Volume vs. surface?}
Purely volumetric rendering models are most amenable to gradient-based optimization and produce excellent view synthesis results \cite{barron2022mipnerf360}. On the other hand, increasing sparsity and moving closer \cite{yariv2021volume, wang2021neus} or completely \cite{chen2022mobilenerf, munkberg2022extracting} to a surface-like representation degrades image quality but results in compact representations that are cheap to render.
\item
\emph{Memory bound vs. compute bound?}
The most compact representations (such as the MLP network in \citet{mildenhall2020nerf} or the low-rank decomposition in \citet{chen2022tensorf}) require many FLOPS to query, and the fastest representations (such as the sparse 3D data structures used in \citet{yu2021plenoctrees} and \citet{hedman2021snerg}) consume large amounts of graphics memory.
\end{itemize}
One approach to this trade-off is to embrace a slower, more compact volumetric model for optimization and to subsequently ``bake'' it into a larger but faster representation for rendering. 
However, baking often affects the representation or rendering model which can lead to a large drop in image quality.
Though this can partially be ameliorated by fine-tuning the baked representation, fine-tuning does not easily scale to larger scenes, as computing gradients for optimization requires significantly more memory than rendering.
 
The goal of our work is to find a representation that is well suited for both optimization and fast rendering. Our solution is a single unified radiance field representation with two different underlying \emph{parameterizations}.
In both stages, our memory-efficient radiance field (MERF) is defined by a combination of a voxel grid~\cite{sun2022direct,yu2021plenoxels} and triplane data structure~\cite{chan2021eg3d}.
During optimization, we use the NGP hash grid structure~\cite{muller2022instant} to compress our parameterization, which allows for differentiable sparsification and provides an inductive bias that aids convergence.
After optimization, we query the recovered NGP to explicitly bake out the MERF and create a binary occupancy grid to accelerate rendering. Critically, both the NGP-parameterized and baked MERF \emph{represent the same underlying radiance field function}. This means that the high quality achieved by the optimized MERF carries over to our real-time browser-based rendering engine.

\section{Related Work}
As our goal is real-time view synthesis in large unbounded scenes, this discussion is focused on approaches that accelerate rendering or reconstruct large spaces. For a comprehensive overview of recent view synthesis approaches, please refer to \citet{tewari2022advances}.

Early methods for real-time large-scale view synthesis either captured a large number of images and interpolated them with optical flow~\cite{aliaga2002sea} or relied heavily on hand-made geometry proxies~\cite{debevec1998efficient,buehler01ulr}. Later techniques used inaccurate, but automatically reconstructed geometry proxies~\cite{chaurasia2013superpixel}, and relied on screen-space neural networks to compensate for this~\cite{hedman2018deep,ruckert2021adop}.
Neural Radiance Fields (NeRF)~\cite{mildenhall2020nerf} facilitated higher quality reconstructions by representing the full scene volume as a multi-layer perceptron (MLP). This volumetric representation can easily model thin structures and semi-transparent objects and is also well-suited to gradient-based optimization.

NeRF was quickly extended to reconstruct large scenes by reconstructing crowdsourced data~\cite{martinbrualla2020nerfw}, tiling the space with NeRF networks~\cite{tancik2022blocknerf,turki2022meganerf}, and reconstructing the scene in a warped domain where far-away regions are compressed~\cite{kaizhang2020,barron2022mipnerf360}.
Later, fast radiance field reconstruction was achieved by representing the scene as a grid, stored either densely~\cite{yu2021plenoxels}, as latent features to be decoded~\cite{sun2022direct,karnewar2022relu}, or as latent hash grids~\cite{muller2022instant} implemented with specialized CUDA kernels~\cite{li2022nerfacc}. While this dramatically reduces reconstruction time, accurate real-time rendering of large scenes has not yet been demonstrated at high resolutions.

Other methods addressed real-time rendering by precomputing and storing (i.e.\ \emph{baking}) NeRF's view-dependent colors and opacities in volumetric data structures~\cite{hedman2021snerg,yu2021plenoctrees,garbin2021fastnerf,zhang2022digging},
or by splitting the scene into voxels and representing each voxel with a small separate MLP~\cite{reiser2021kilonerf}. However, these representations consume a lot of graphics memory and are thus limited to objects, not scenes. Furthermore, these methods incur a quality loss during baking due to the mismatch between the slower rendering procedure used for training and the real-time rendering procedure used for inference.

Alternatively, faster rendering can be achieved by extending the network to work with ray segments rather than points~\cite{lindell2021autoint, wu2022diver} or by training a separate sampling network~\cite{barron2022mipnerf360,neff2021donerf,piala2021terminerf,kurz2022adanerf}. However, these approaches have not achieved real-time rates at high resolutions, likely because they require evaluating an MLP for each sample along a ray. Light field coordinates circumvent this problem and require just one MLP evaluation per ray~\cite{sitzmann2021lfns,li2022neulf,wang2022r2l,attal2022learning}. However, like traditional light field representations~\cite{gortler1996lumigraph,levoy1996light}, this approach has only been demonstrated to work well within small viewing volumes. Similarly, multi-plane image~\cite{wizadwongsa2021nex,flynn2019deepview,zhou2018stereomag,mildenhall2019llff} or multi-sphere image~\cite{broxton20,attal2020matryodska} representations map well to graphics hardware and can be rendered in real-time, but also support only restricted camera motion.

It is also possible to speed-up NeRF rendering by post-processing the output image with a convolutional neural network. This makes it possible to perform an expensive volumetric rendering step at a lower resolution and then upsample that result to the final desired resolution~\cite{li2022steernerf,wang20224knerf}. \citet{wu2022snisr} combined this approach with baking and showed high-quality real-time rendering of large scenes. However, to achieve this, they required a 3D scan of the scene as input, and they used a CUDA implementation designed for workstation-class hardware. In contrast, our method only needs posed images as input and runs in a browser on commodity hardware such as laptops.

Recent methods achieve extremely fast rendering by constraining the NeRF network evaluations to planes or polygons~\cite{chen2022mobilenerf,lin2022neurmips}. While this works well for objects or limited camera motion, we show that these approaches introduce a loss in quality for large unbounded scenes.

The problem of compressing NeRF reconstructions has also been explored in prior work. Several methods achieve this by post-processing an existing reconstruction through incremental pruning~\cite{deng2023compressing} with vector quantization~\cite{li2022compressing}. \citet{takikawa2022variable} directly optimize for a compressed codebook-based representation of the scene. While these methods all report impressive compression ratios, they all rely on evaluating an MLP for each volume sample and are therefore too slow for real-time rendering of large scenes.

Our approach works by projecting 3D samples onto three 2D projections that correspond to the cardinal axes. Similar representations, often referred to as \emph{tri-planes}, have been explored for surface reconstruction from point clouds~\cite{peng2020convolutional} and generative modelling of 3D scenes~\cite{devries2021unconstrained} or faces~\cite{chan2021eg3d}. Recently TensoRF~\cite{chen2022tensorf} use tri-planes for NeRF reconstruction. TensorRF decomposes the 3D scene volume into a sum of vector-matrix outer products, which makes it possible to directly train a compressed and high quality radiance field. However, TensoRF trades off memory footprint for more expensive queries that involve a large matrix multiplication. Our representation significantly speeds up the query time by removing the need for the matrix product while simultaneously halving the memory bandwidth consumption.

\section{Preliminaries}\label{sec:bg} 
We begin with a short review of relevant prior work on radiance fields for unbounded scenes. A radiance field maps every 3D position $\mathbf{x}\in\mathbb{R}^3$ and viewing direction $\mathbf{d}\in\mathbb{S}^2$ to the volumetric density $\density\in\mathbb{R}_+$ at that location and the RGB color emitted from it along the view direction, $\mathbf{c}\in\mathbb{R}^3$. The color of the ray emitted from point $\mathbf{o}$ in the direction $\mathbf{d}$ can then be computed using the radiance field by sampling points along the ray, $\mathbf{x}_i = \mathbf{o} + t_i\mathbf{d}$, and compositing the corresponding densities $\{\density_i\}$ and colors $\{\mathbf{c}_i\}$ according to the numerical quadrature approach of~\citet{max1995optical}:

\begin{equation}
    \mathbf{C} = \sum_{i} w_i \mathbf{c}_i\,, \,\,\,
    w_i = \alpha_i T_i \,, \,\,\,
    T_i=\prod_{j=1}^{i-1} (1 - \alpha_j)\,, \,\,\,
    \alpha_i = 1 - e^{-\density_i \delta_i}\,, \label{eq:vol}
\end{equation}
where $T_i$ and $\alpha_i$ denote transmittance and alpha values of sample $i$, and $\delta_i = t_{i+1}-t_i$ is the distance between adjacent samples. 

The original NeRF work parameterized a radiance field using a Multilayer Perceptron (MLP), which outputs the volume density and view-dependent color for any continuous 3D location. In order to reduce the number of MLP evaluations to one per ray, SNeRG uses a deferred shading model in which the radiance field is decomposed into a 3D field of densities $\density$, diffuse RGB colors $\colordiffuse$, and feature vectors $\mathbf{f}$~\cite{hedman2021snerg}.

SNeRG's deferred rendering model volumetrically accumulates the diffuse colors $\{\mathbf{c}_{d,i}\}$ and features $\{\mathbf{f}_i\}$ along the ray, similar to Equation~\ref{eq:vol}:
\begin{equation}
    \mathbf{C}_d = \sum_{i} w_i \mathbf{c}_{d,i}, \quad \mathbf{F} = \sum_{i} w_i \mathbf{f}_{i}, \label{eq:snergvol}
\end{equation}
and computes the ray's color as the sum of the accumulated diffuse color $\mathbf{C}_d$ and the view-dependent color computed using a small MLP $h$ that takes as input $\mathbf{C}_d$, $\mathbf{F}$, and the viewing direction $\mathbf{d}$:

\begin{equation}
    \mathbf{C} = \mathbf{C}_d + h(\mathbf{C}_d, \mathbf{F}, \mathbf{d}).
\end{equation}
SNeRG uses a large MLP during training and bakes it after convergence into a block-sparse grid for real-time rendering.

In order for radiance fields to render high quality unbounded scenes containing nearby objects as well as objects far from the camera, mip-NeRF 360~\cite{barron2022mipnerf360} uses a contraction function to warp the unbounded scene domain into a finite sphere:
\begin{equation}
\label{eqn:contract360}
\operatorname{contract}(\mathbf{x}) = \begin{cases}
\mathbf{x} &\text{if $\norm{\mathbf{x}}_2 \leq 1$} \\
\left(2 - \frac{1}{\prescript{}{\phantom{2}}{\norm{\mathbf{x}}}_2}\right)\frac{\mathbf{x}}{\prescript{}{\phantom{2}}{\norm{\mathbf{x}}}_2} & \text{if $\norm{\mathbf{x}}_2 > 1$}
\end{cases}
\end{equation}

\begin{figure}[t]
  \centering
  \includegraphics[width=\linewidth]{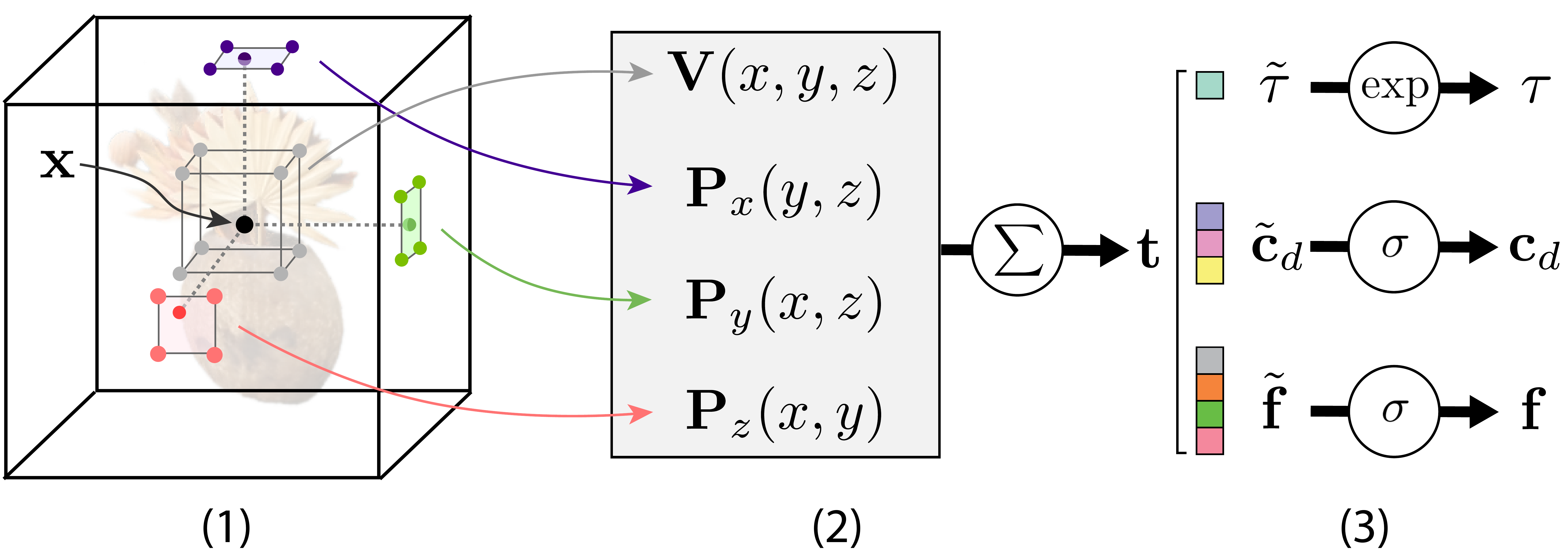}
  \caption{Our scene representation. For a location $\mathbf{x}$ along a ray: (1) We query its eight neighbors on a low-resolution 3D grid; and we project it onto each of the three axis-aligned planes, and then query each projection’s four neighbors on a high-resolution 2D grid. (2) The eight low-resolution 3D neighbors are evaluated and trilinearly interpolated while the three sets of four high-resolution 2D neighbors are evaluated and bilinearly interpolated, and the resulting features are summed into a single feature vector $\mathbf{t}$. (3) The feature vector is split and nonlinearly mapped into three components: density $\density$, RGB color $\mathbf{c}_d$, and a feature vector $\mathbf{f}$ encoding view dependence effects.}
  \label{fig:param}  
\end{figure}

\section{Scene Representation}

In this section, we describe the MERF scene representation, which is designed to enable real-time volumetric rendering of unbounded scenes while maintaining a low memory footprint. 

\subsection{Volume Parameterization}

MERF represents a scene using a 3D field of volume densities $\density \in \mathbb{R}_+$, diffuse RGB colors $\colordiffuse\in\mathbb{R}^3$, and feature vectors $\mathbf{f}\in\mathbb{R}^K$, as shown in Figure~\ref{fig:param}. These quantities are rendered using the deferred shading model from SNeRG, described in Section~\ref{sec:bg}. 

We parameterize this field with a low-resolution 3D $L\times L\times L$ voxel grid $V$ and three high-resolution 2D $R\times R$ grids $P_x$, $P_y$, and $P_z$, one for each of the cardinal $yz$, $xz$, and $xy$ planes.
Each element of the low-resolution 3D grid and the three high-resolution 2D grids stores a vector with $C = 4+K$ channels 
In our experiments, we use $C=8$ and default to $L=512$ and $R=2048$.

We define the continuous field of $C$-vectors as the sum of trilinearly interpolated vectors from the 3D grid and bilinearly interpolated vectors from the three 2D grids:
\begin{equation}
\mathbf{t}(x, y, z) = \mathbf{V}(x, y, z) + \mathbf{P}_x(y, z) + \mathbf{P}_y(x, z) + \mathbf{P}_z(x, y),
\end{equation}
where $\mathbf{V}\colon\mathbb{R}^3\rightarrow\mathbb{R}^C$ is a trilinear interpolation operator using the 3D grid values, and $\mathbf{P}_i\colon\mathbb{R}^2\rightarrow\mathbb{R}^C$ is a bilinear interpolation operator using the grid perpendicular to the $i$th axis, for $i\in \{x, y, z\}$.

We split the $C$-vector at any 3D location into three components corresponding to density $\tilde{\density}\in\mathbb{R}$, diffuse color $\tilde{\mathbf{c}}_d\in\mathbb{R}^3$, and view-dependence feature $\tilde{\mathbf{f}}\in\mathbb{R}^K$, and then apply nonlinear functions to obtain the three values:
\begin{equation} \label{eq:nonlinearities}
    \density = \exp(\tilde{\density}), \quad \colordiffuse = \sigma(\tilde{\mathbf{c}}_d),  \quad \mathbf{f} = \sigma(\tilde{\mathbf{f}}),
\end{equation}
where $\sigma$ is the standard logistic sigmoid function, which constrains colors and features to lie within $(0, 1)$. Note that we apply the nonlinearities after interpolation and summation, which has been shown to greatly increase the representational power of grid representations~\cite{karnewar2022relu, sun2022direct}.

\subsection{Piecewise-projective Contraction}
\label{sec:contraction}

Mip-NeRF 360~\cite{barron2022mipnerf360} demonstrated the importance of applying a contraction function to input coordinates when representing large-scale scenes with unbounded extent. The contraction maps large far-away regions of space into small regions in contracted space, which has the effect of allocating model capacity towards representing high-resolution content near the input camera locations.

\begin{figure}[t!]
  \centering
     \begin{subfigure}[b]{0.48\linewidth}
         \centering
          \includegraphics[width=\textwidth]{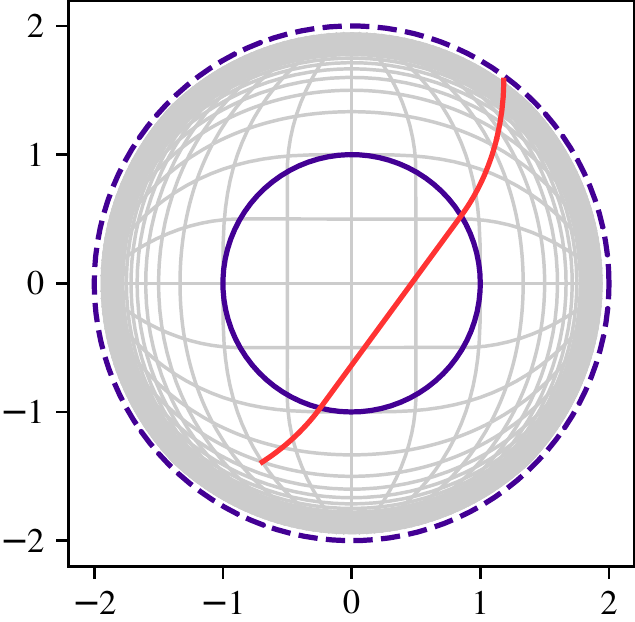}
         \caption{$\operatorname{contract}(\cdot)$}
         \label{fig:contractsphere}
     \end{subfigure}
     \begin{subfigure}[b]{0.48\linewidth}
         \centering
          \includegraphics[width=\textwidth]{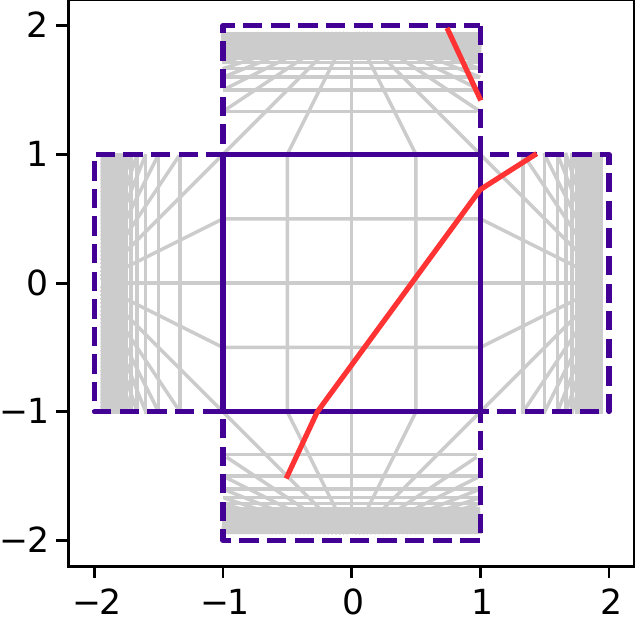}
         \caption{$\newcontract(\cdot)$}
         \label{fig:contractndc}
     \end{subfigure}
  \caption{A 2D visualization of (a) the spherical contraction function used by \citet{barron2022mipnerf360}; and (b) our piecewise projective contraction, both applied to the same ray (in red). The spherical contraction maps a straight line to a curve, which makes empty space skipping more expensive, while our contraction maps a straight line to a small number of line segments. The gray lines show the result of applying each contraction function to a regular grid.}
  \label{fig:contract}
\end{figure}

The contraction function used in mip-NeRF 360 (Equation~\ref{eqn:contract360}) nonlinearly maps any point in space $\mathbf{x}\in\mathbb{R}^3$ into a radius-$2$ ball, and represents the scene within this contracted space. While mip-NeRF 360's contraction function is effective and efficient to evaluate, it cannot be easily used in a real-time rendering pipeline with discretized voxels. This is because empty space skipping is critical for efficient volume rendering, and requires a method for analytically computing the intersection between a ray and an axis-aligned bounding box (AABB) of any ``active'' (\ie, occupied) content. As can be seen in Figure~\ref{fig:contractsphere}, mip-NeRF 360's contraction function does not preserve straight lines, and thereby makes ray-AABB intersections challenging to compute. 

To address this, we propose a novel contraction function for which ray-AABB intersections can be computed trivially. The $j$th coordinate of a contracted point is defined as follows:
\begin{equation}
\newcontract(\mathbf{x})_j =\begin{cases}
x_j &\text{if } \|\mathbf{x}\|_\infty \leq 1\\
\frac{x_j}{\prescript{}{\phantom{j}}{\|\mathbf{x}\|_\infty}} &\text{if } x_j \neq \|\mathbf{x}\|_\infty > 1 \\
\left(2 - \frac{1}{|x_j|}\right) \frac{x_j}{|x_j|} & \text{if } x_j = \|\mathbf{x}\|_\infty > 1\end{cases} \, ,
\end{equation}
where $\|\cdot\|_\infty$ is the $L_\infty$ norm ($\|\mathbf{x}\|_\infty = \max_j |x_j|$). This contraction function is piecewise-projective: it contains seven regions, and within each region, it is a projective transformation. The unit cube $\|\mathbf{x}\|_\infty$ is preserved (\ie, the contraction function is the identity), and the other six regions defined by the coordinate maximizing $|x_j|$ and its sign each get mapped by a different projective transformation. Because projective transformations preserve straight lines, any contracted ray must be piecewise-linear, and the only discontinuities in its direction are on the boundaries between the seven regions (see Figure~\ref{fig:contractndc} for a 2D example with 5 regions). The origin and direction of a ray can therefore be computed in contracted space, which allows us to use standard ray-AABB intersection tests. Table~\ref{tab:model_analysis} (c) shows that our proposed contraction function performs on par with the original spherical contraction.

\section{Training and Baking}

In this section, we describe how to efficiently optimize a MERF and bake it into a representation that can be used for high-quality real-time rendering. To achieve our goal of not losing quality during this baking process, we take care to ensure that the baked representation and the scene representation we use during optimization both describe the same radiance field. This requires a training procedure that accounts for any baking discretization or quantization.

\subsection{Efficient Training}

Modeling high-resolution large-scale scenes requires high-capacity representations that may consume prohibitive amounts of memory during training. Significantly more memory is consumed during training than during rendering because training requires that intermediate activations be stored for the sake of backpropagation and that higher-precision per-parameter statistics be accumulated for the Adam optimizer. In our case, we found that training requires more than twelve times as much video memory as rendering. We thus optimize a compressed representation of MERF's low-resolution voxel grid and high-resolution 2D planes by parameterizing them as MLPs with a multi-resolution hash encoding~\cite{muller2022instant}.

However, baking the MLPs' outputs onto discrete grids for rendering introduces a mismatch between the representations used for training and for rendering. Prior work accounted for this by fine-tuning the baked representation \cite{yu2021plenoxels}, but fine-tuning requires the entire representation to be stored in memory and suffers from the aforementioned scalability issues. Instead, we simulate finite grid resolutions during training by querying the MLPs at virtual grid corners and interpolating the resulting outputs using bilinear interpolation for the high-resolution 2D grids and trilinear interpolation for the low-resolution voxel grid.

In addition to requiring high-capacity representations, high-resolution large-scale scenes require dense samples along rays during volume rendering, which also significantly contributes to the training memory footprint. To address this, mip-NeRF 360 introduced a hierarchical sampling technique that uses ``proposal'' MLPs to represent coarse versions of scene geometry. A proposal MLP maps 3D positions to density values, which are converted into probability distributions along rays that are supervised to be consistent with the densities output by the NeRF MLP. These proposal distributions are used in an iterative resampling procedure that produces a small number of samples that are concentrated around visible scene content. While this proposal MLP hierarchical sampling strategy is effective for reducing the number of samples along each ray during training, the proposal MLP is too expensive to evaluate for real-time rendering purposes.

Instead, we use traditional empty space skipping during rendering, which also concentrates representation queries around surfaces. To avoid introducing a mismatch between training and rendering, we only bake content in regions considered occupied by the proposal MLP during training, as detailed in Section~\ref{sec:baking}. 

\subsection{Quantization}
To reduce our system's memory consumption at render time, we wish to quantize each of the $C$ dimensions at every location in the grid to a single byte (see further discussion in Section~\ref{sec:rendering}). However, simply quantizing the optimized grid values after training creates mismatches between the optimized model and the one used for rendering, which leads to a drop in rendering quality as shown in Table~\ref{tab:model_analysis} (b).

Our solution to this is to quantize the $C$ values at every location during optimization. That is, we nonlinearly map them to lie in $[0, 1]$ using a sigmoid $\sigma$, then quantize them to a single byte using a quantization function $q$, and finally affinely map the result to the range $[-m, m]$, as:
\begin{equation}
    \tilde{\mathbf{t}}' = 2m\cdot q\!\left(\sigma\!\left(\tilde{\mathbf{t}}\right)\right) - m \, ,
\end{equation}
where we choose $m=14$ for densities (which are computed using an exponential nonlinearity), and $m=7$ for diffuse colors and features. Note that this only quantizes the values stored in the grid, and the non-linearities in Equation~\ref{eq:nonlinearities} are subsequently applied after linearly interpolating and summing these values.

We implement the byte quantization function $q$ as:
\begin{equation}
    q(x) = x + \cancel{\nabla}\!\left(\frac{\lfloor (2^8-1) x + \sfrac{1}{2} \rfloor}{2^8-1} - x\right)  \, ,
\end{equation}
where $\cancel{\nabla}(\cdot)$ is a stop-gradient, which prevents gradients from backpropagating to its input. This use of a stop-gradient allows us to obtain gradients for the non-differentiable rounding function by treating $q$ as the identity function during the backward pass, which is referred to as the straight-through estimator, as used in~\cite{bengio2013estimating,yin2019understanding}.

\subsection{Baking}
\label{sec:baking}
After training, we evaluate and store the MLP's outputs on discrete grids for real-time rendering. First, we compute a binary 3D grid $\textbf{A}$ indicating voxels that contributed to any training image (\ie, voxels should \emph{not} be stored if they correspond to occluded content, are not sampled by any training ray, or have low opacity). To populate $\textbf{A}$, we render all training rays and extract from them a set of weighted points $\{(\mathbf{x}_i, w_i)\}$, where $\mathbf{x}_i$ is the point's position, and $w_i$ is the associated volume rendering weight from Equation~\ref{eq:vol}. Note that these points cluster around surfaces in the scene as they are sampled with a proposal-MLP~\cite{barron2022mipnerf360}.

We mark the eight voxels surrounding a given point $\mathbf{x}_i$ as occupied if both the volume rendering weight $w_i$ and the opacity, $\alpha_i$, exceed a threshold set to $0.005$. To cull as aggressively as possible, we compute $\alpha_i$ based on the distance between samples $\delta_i$ used by the real-time renderer --- recall that it steps through contracted space with a small uniform step size. As the proposal-MLP often suggests steps larger than $\delta_i$, computing $\alpha_i$ this way leads to better culling. However, we still guarantee voxels which contribute a significant opacity value ($\alpha_i > 0.005$) are not culled in the final sampling scheme. Note that while the opacity $\alpha_i$ only depends on the density at $\mathbf{x}_i$, the weight $w_i$ also depends on densities along the entire ray, making usage of $w_i$ necessary to account for visibility.
We observe that the opacity check based on the real-time renderer's step size significantly decreases the fraction of the volume marked as occupied. 
Note that this is in addition to the sparsity already achieved by only considering voxels in locations that have been sampled by the proposal-MLP.
In contrast, existing baking pipelines often do not consider the proposal-MLP and perform visibility culling with uniformly-spaced sample points. This often results in fog-like artifacts and floating blobs because the underlying 3D field can have arbitrary values in regions not sampled by the proposal-MLP. Table~\ref{tab:model_analysis} demonstrates that our Proposal-MLP-aware baking pipeline is almost lossless.\\
After computing the binary grid $\textbf{A}$, we bake the three high-resolution 2D planes and the low-resolution 3D voxel grid. Following SNeRG, we store this voxel grid in a block-sparse format, where we only store data blocks that contain occupied voxels.
For empty space skipping, we create multiple lower resolution versions of the binary occupancy grid $\textbf{A}$ with max-pooling. To reduce storage, we encode textures as PNGs.

\newcommand{\realtimesymbol}{$\color{gold} \star$}

\definecolor{realtimecolor}{rgb}{1.,1,0.85}
\sethlcolor{realtimecolor}

\begin{table*}[t!]
\centering
\begin{tabular}{l|ccc|ccc}
\multicolumn{1}{c}{\phantom{x}} & \multicolumn{3}{c}{Outdoor} & \multicolumn{3}{c}{Indoor} \\
& PSNR $\uparrow$ & SSIM $\uparrow$ & LPIPS $\downarrow$ & PSNR $\uparrow$ & SSIM $\uparrow$ & LPIPS $\downarrow$ \\ \hline \hline
NeRF \cite{mildenhall2020nerf}                 &                   21.46 &                   0.458 &                   0.515 &                   26.84 &                   0.790 &                   0.370 \\
NeRF++ \cite{kaizhang2020}                     &                   22.76 &                   0.548 &                   0.427 &                   28.05 &                   0.836 &                   0.309 \\
SVS~\cite{riegler2021stable} & 23.01 & 0.662 &    \bf{0.253} & 28.22 & 0.907 &    \bf{0.160} \\
Mip-NeRF 360 \cite{barron2022mipnerf360}       &    \bf{24.47} &    \bf{0.691} & 0.283 &    \bf{31.72} &    \bf{0.917} & 0.180 \\
Instant-NGP \cite{muller2022instant}           &                   22.90 &                   0.566 &                   0.371  & 29.15 & 0.880 & 0.216 \\ \hline
\rowcolor{realtimecolor} Deep Blending \cite{hedman2018deep}            &                   21.54 &                   0.524 &                   0.364 &                   26.40 &                   0.844 &                   0.261 \\
\rowcolor{realtimecolor} Mobile-NeRF \cite{chen2022mobilenerf}          &                   21.95 &                   0.470 &                   0.470 & $-$ & $-$ & $-$ \\
\rowcolor{realtimecolor} Ours                                           & \bf{23.19} & \bf{0.616} & \bf{0.343}  & \bf{27.80} & \bf{0.855} & \bf{0.271}
\end{tabular}
\caption{
Quantitative results of our model on all scenes from Mip-NeRF 360~\cite{barron2022mipnerf360}. Models that render in real-time are \hl{highlighted}. Mobile-NeRF did not evaluate on the  ``indoor'' scenes, so those metrics are absent.
}
\label{tab:alldoor}
\end{table*}

\section{Real-time Rendering} \label{sec:rendering}

We implement our real-time viewer as a Javascript 3D (three.js) web application, based on SNeRG's implementation, where rendering is orchestrated by a single GLSL fragment shader. 

For efficient ray marching, we employ a multi-resolution hierarchy of occupancy grids. The set of occupancy grids is created by max-pooling the full-resolution binary mask $\textbf{A}$ with filter sizes $16, 32$ and $128$. For instance, if the base resolution is $4096$, this results in occupancy grids of size $256, 128$ and $32$, occupying a total of 18 MB of video memory. 
We leverage this multi-resolution hierarchy of occupancy grids for faster space skipping. Given any sample location, we query the occupancy grids in a coarse-to-fine manner. If any level indicates the voxel as empty, we can skip the corresponding volume until the ray enters a new voxel at that level and compute the new sample location using the efficient ray-AABB intersection discussed in Section~\ref{sec:contraction}. We only access the MERF scene representation for samples where all occupancy grid levels are marked as occupied.
Finally, we terminate ray marching along a ray once the transmittance value $T_i$ (defined in Equation~\ref{eq:vol}) falls below $2 \times 10^{-4}$.

To further decrease the number of memory accesses during rendering, we split textures into density and appearance (containing diffuse RGB colors and feature vectors) components. When accessing the MERF representation at any location, we first query the density component and only read the corresponding appearance component if the voxel opacity computed from the returned density is nonzero.
Moreover, we obtain an additional $4\times$ speed-up by optimizing the deferred rendering MLP. More specifically, we conduct loop unrolling, modify the memory layout to facilitate linear accesses, and exploit fast \textit{mat4}-multiplication. 

\section{Experiments}
We experimentally evaluate our model in terms of rendering quality, video memory consumption, and real-time rendering performance. We compare MERF to a variety of offline view synthesis methods (NeRF \cite{mildenhall2020nerf}, mip-NeRF 360 \cite{barron2022mipnerf360}, Stable View Synthesis \cite{riegler2021stable}, and Instant-NGP~\cite{muller2022instant}), and real-time ones (Deep Blending \cite{hedman2018deep}, Mobile-NeRF \cite{chen2022mobilenerf}, and SNeRG \cite{hedman2021snerg}). To make this evaluation as rigorous as possible we evaluate against an improved version of SNeRG, which we call SNeRG++. SNeRG++ uses many components of our approach: multi-level empty space skipping, an optimized MLP implementation, our improved baking pipeline, and post-activation interpolation (which increases model expressivity by allowing for intra-voxel discontinuities~\cite{sun2022direct, karnewar2022relu}). Unless otherwise stated, for MERF we set the triplane resolution $R$ to 2048 and the sparse grid resolution $L$ to 512, and for SNeRG++ we set the grid resolution to $2048$.

For evaluation, we use the challenging mip-NeRF 360 dataset \cite{barron2022mipnerf360} which contains five outdoor and four indoor scenes. All scenes are unbounded and require a high resolution representation ($2048^3$) to be faithfully reproduced. If not indicated otherwise, reported metrics are averaged over runs from the five outdoor scenes. We evaluate rendering quality using peak-signal-to-noise-ratio (PSNR), SSIM~\cite{wang2004image}, and LPIPS~\cite{zhang2018lpips}.

\subsection{Quality Comparison}
We first compare our method to a variety of offline and real-time view synthesis methods in terms of rendering quality on both outdoor and indoor scenes. As can be seen in Table~\ref{tab:alldoor}, MERF achieves better scores than the real-time competitors DeepBlending and Mobile-NeRF on all three metrics. On the outdoor scenes, MERF even achieves higher scores than the offline methods Instant-NGP, NeRF and NeRF++ and performs on par with SVS in terms of PSNR and SSIM, despite offline rendering quality not being the primary focus of this work. In contrast, on indoor scenes MERF achieves slightly lower scores than Instant-NGP, SVS, and NeRF++. This might be a result of the indoor scenes containing more artificial materials with stronger view-dependent effects that our shallow decoder MLP $h$
designed for real-time rendering cannot model well. Qualitatively, Figure~\ref{fig:visual_comparison} shows that our reconstructions of the background are much sharper compared to Mobile-NeRF and Instant-NGP, which we attribute to using a scene contraction function.

\subsection{MERF vs. SNeRG++}
\begin{figure}[t!]
  \centering
  \includegraphics[width=\linewidth]{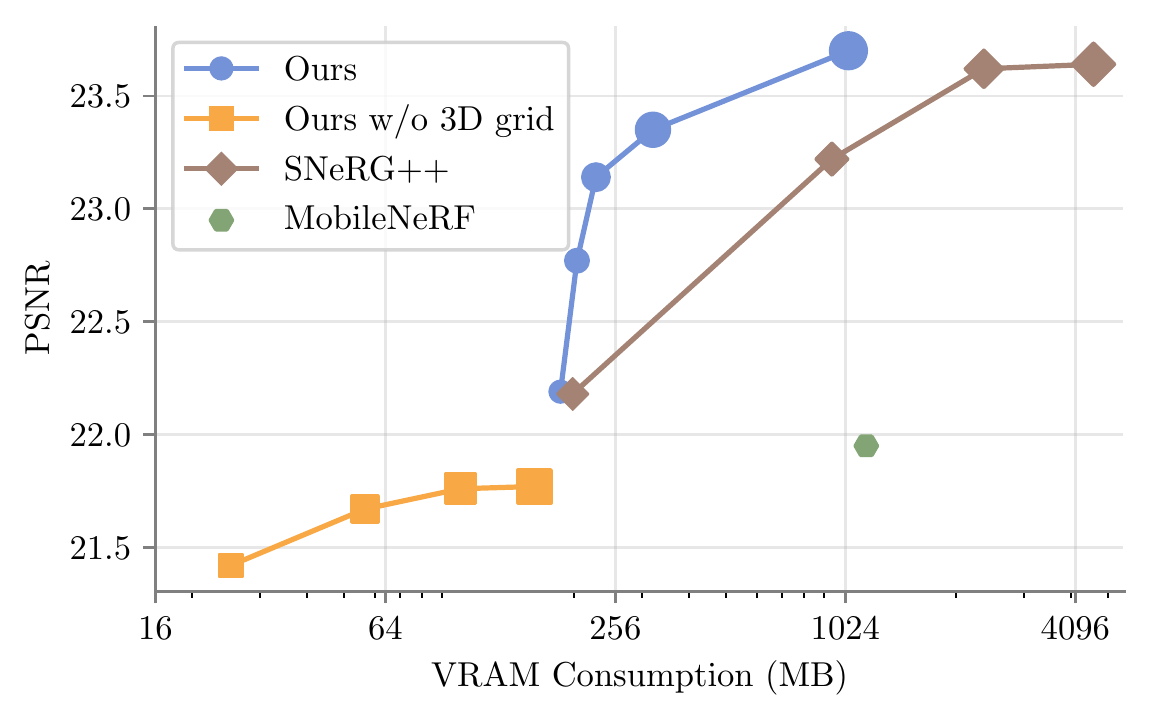}
  \vspace{-0.3in}
  \caption{PSNR (higher is better) vs VRAM consumption (lower is better) for our model, our improved SNeRG baseline, MobileNeRF, and an ablation of our model without 3D grids. Each line besides MobileNeRF represents the same model with a varying resolution of its underlying spatial grid, indicated by marker size.}
  \label{fig:vram_vs_psnr}
\end{figure}

\begin{figure*}[h]
\centering
\resizebox{0.99\linewidth}{!}{
\begin{tabular}[!t]{@{}c@{}c@{}}

\makecell[c]{
  \includegraphics[width=0.2\linewidth]{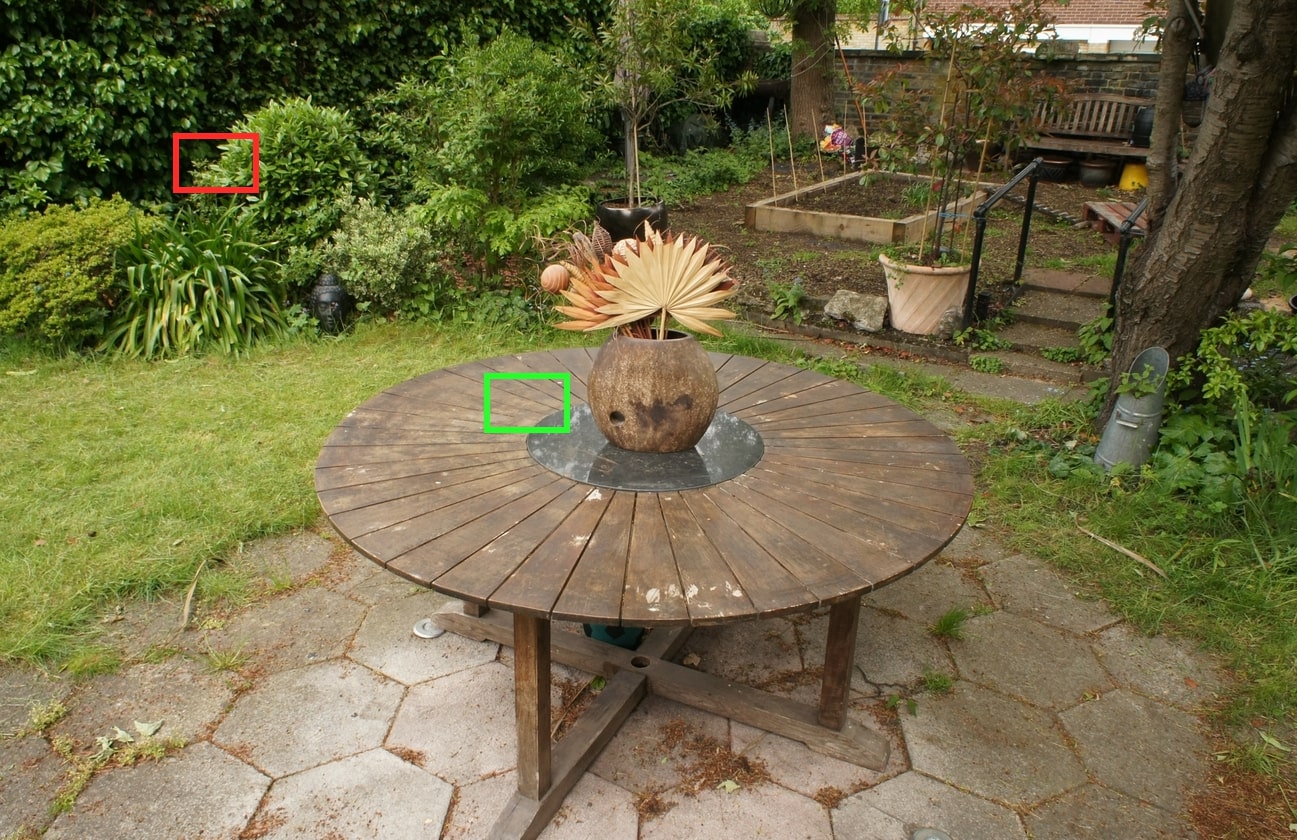}\\
  \texttt{gardenvase}
  } & 
  \begin{tabular}[t]{@{\,}c@{\,}c@{\,}c@{\,}c@{\,}c}
  \includegraphics[width=0.15\linewidth]{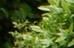} & 
  \includegraphics[width=0.15\linewidth]{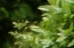} & 
  \includegraphics[width=0.15\linewidth]{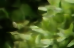} & 
  \includegraphics[width=0.15\linewidth]{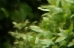} & 
  \includegraphics[width=0.15\linewidth]{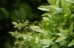} \\
  \includegraphics[width=0.15\linewidth]{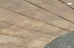} & 
  \includegraphics[width=0.15\linewidth]{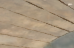} & 
  \includegraphics[width=0.15\linewidth]{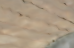} & 
  \includegraphics[width=0.15\linewidth]{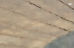} & 
  \includegraphics[width=0.15\linewidth]
  {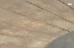} \\
  Ground truth & Ours (2048/512) & SNeRG++ (512) & SNeRG++ (1024) & SNeRG++ (2048) \\
   & 198 MB & 117 MB & 489 MB & 2372 MB
  \end{tabular}\vspace{.5em} \\

\makecell[c]{
  \includegraphics[width=0.2\linewidth]{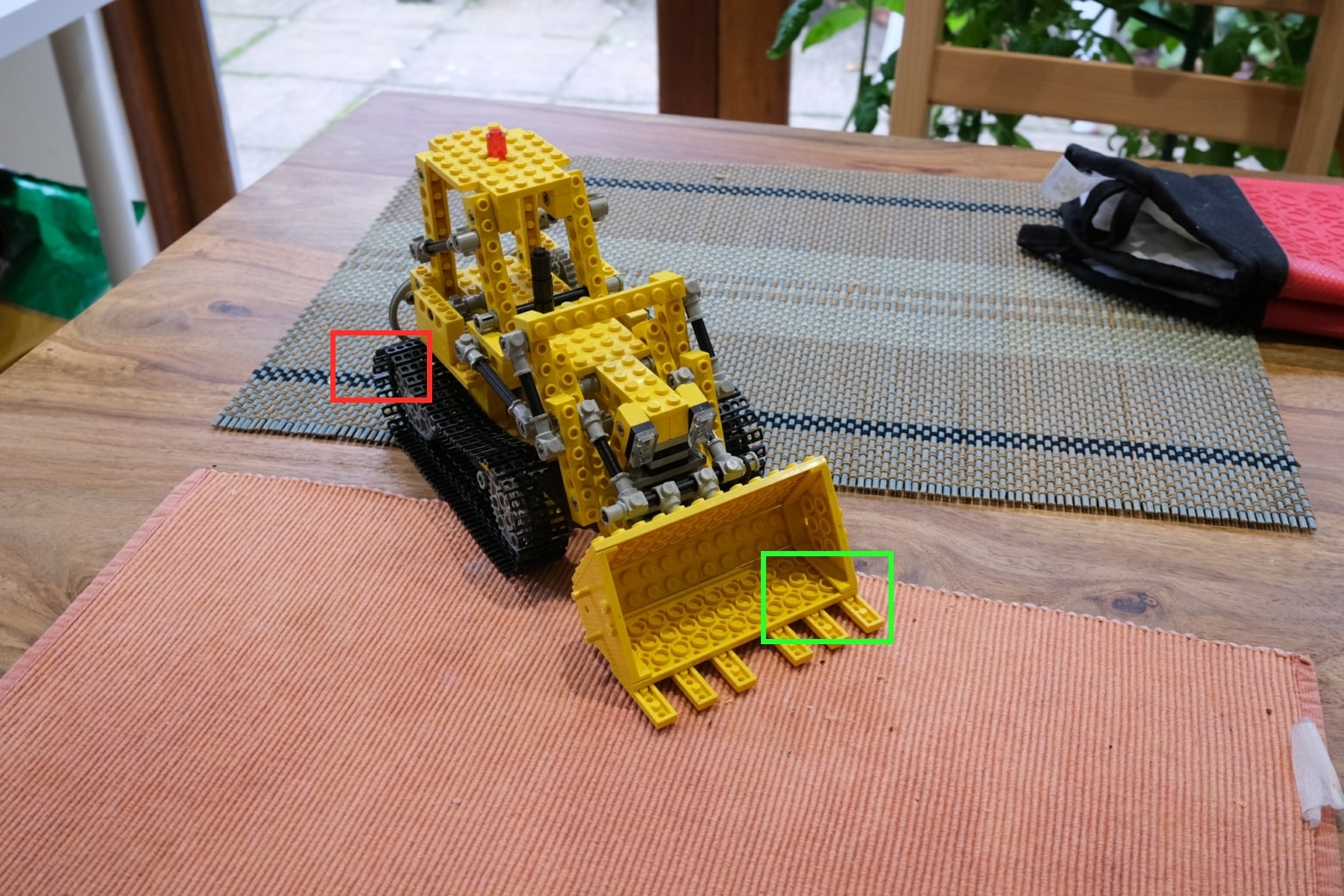}\\
  \texttt{kitchenlego}
  } & 
  \begin{tabular}[t]{@{\,}c@{\,}c@{\,}c@{\,}c@{\,}c}
  \includegraphics[width=0.15\linewidth]{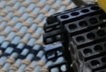} & 
  \includegraphics[width=0.15\linewidth]{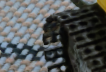} & 
  \includegraphics[width=0.15\linewidth]{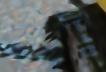} & 
  \includegraphics[width=0.15\linewidth]{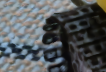} & 
  \includegraphics[width=0.15\linewidth]{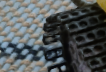} \\
  \includegraphics[width=0.15\linewidth]{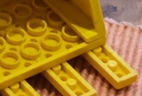} & 
  \includegraphics[width=0.15\linewidth]{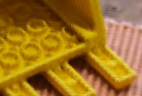} & 
  \includegraphics[width=0.15\linewidth]{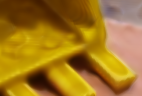} & 
  \includegraphics[width=0.15\linewidth]{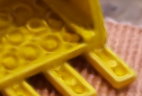} & 
  \includegraphics[width=0.15\linewidth]
  {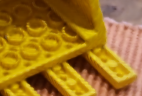} \\
  Ground truth & Ours (2048/512) & SNeRG++ (512) & SNeRG++ (1024) & SNeRG++ (2048) \\
  & 233 MB & 213 MB & 1000 MB & 4570 MB
  \end{tabular}\vspace{.5em} \\

\makecell[c]{
  \includegraphics[width=0.2\linewidth]{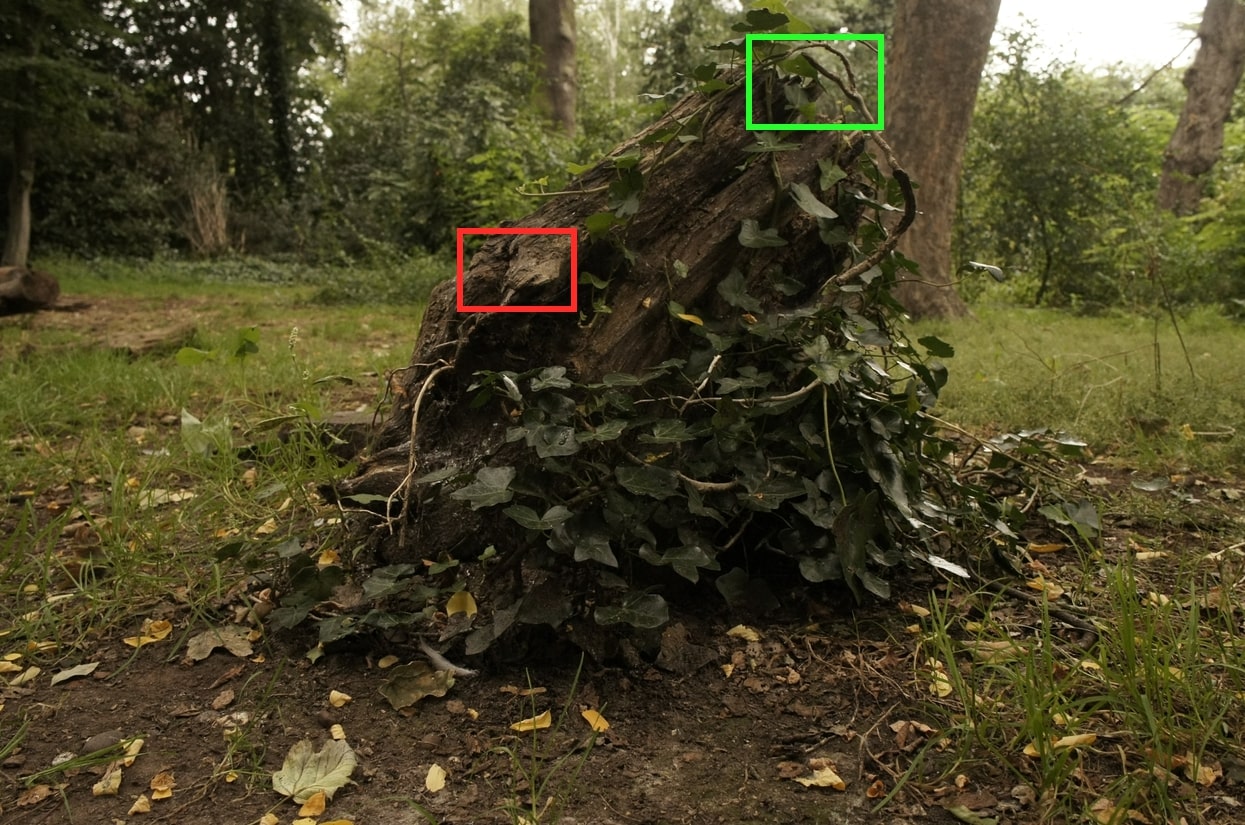}\\
  \texttt{stump}
  } & 
  \begin{tabular}[t]{@{\,}c@{\,}c@{\,}c@{\,}c@{\,}c}
  \includegraphics[width=0.15\linewidth]{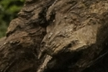} & 
  \includegraphics[width=0.15\linewidth]{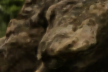} & 
  \includegraphics[width=0.15\linewidth]{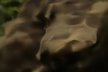} & 
  \includegraphics[width=0.15\linewidth]{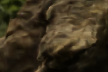} & 
  \includegraphics[width=0.15\linewidth]{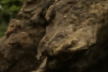} \\
  \includegraphics[width=0.15\linewidth]{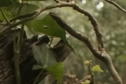} & 
  \includegraphics[width=0.15\linewidth]{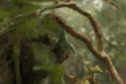} & 
  \includegraphics[width=0.15\linewidth]{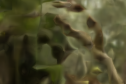} & 
  \includegraphics[width=0.15\linewidth]{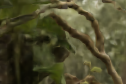} & 
  \includegraphics[width=0.15\linewidth]
  {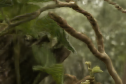} \\
  Ground truth & Ours (2048/512) & SNeRG++ (512) & SNeRG++ (1024) & SNeRG++ (2048) \\
  & 228 MB & 210 MB & 810 MB & 3904 MB
  \end{tabular} \\

\end{tabular}
}
\caption{Visual comparison between MERF and various resolution SNeRG++~\cite{hedman2021snerg} models. Total VRAM (GPU memory) usage during rendering is listed beneath each method name. Only SNeRG++ (512) has comparable size to our model, whereas SNeRG++ (1024) and SNeRG++ (2048) are significantly larger.}
\label{fig:visual_comparison_snerg}
\end{figure*}

\begin{figure*}[t]
\centering
\resizebox{0.99\linewidth}{!}{
\begin{tabular}[!t]{@{}c@{}c@{}c@{}c@{}}

\makecell[c]{\includegraphics[width=0.33\linewidth]{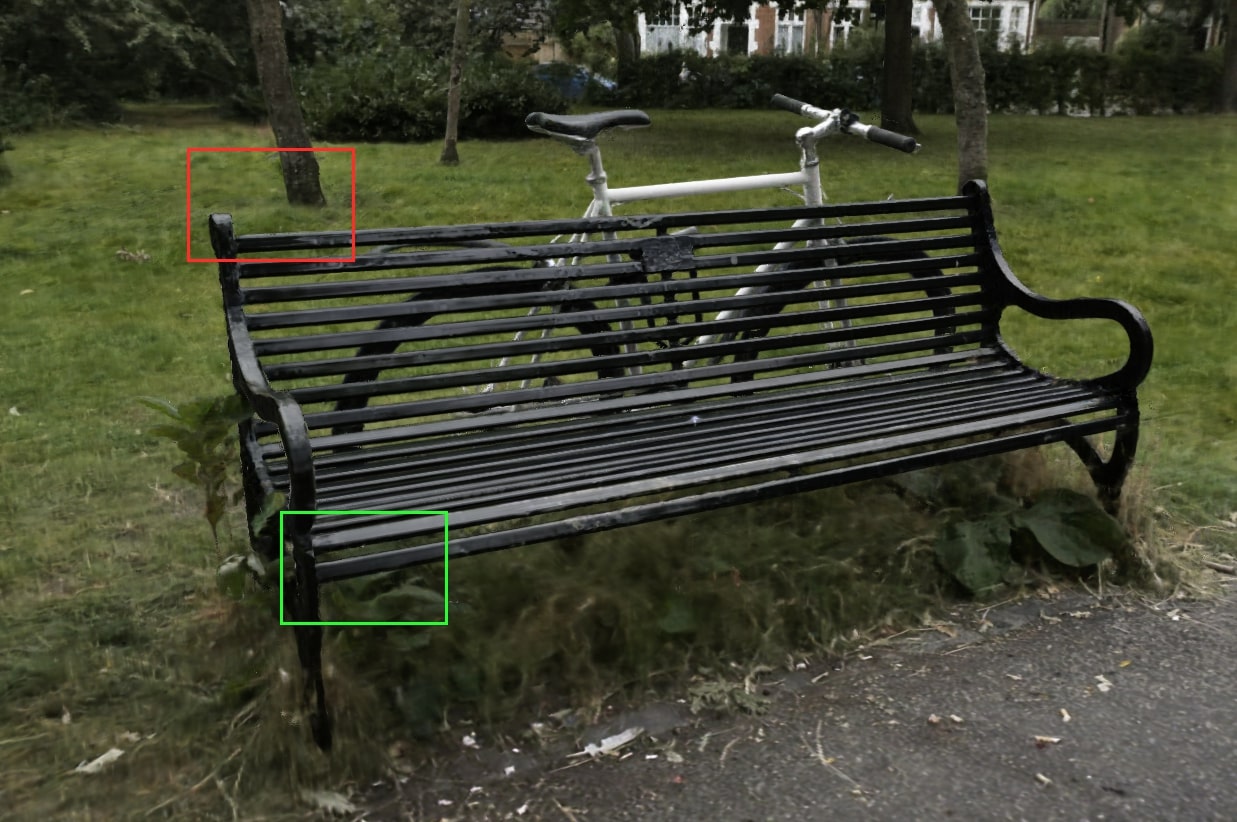}}

&
  
\makecell[c]{
  \begin{tabular}[t]{@{\,}c}
  \includegraphics[width=0.16\linewidth]{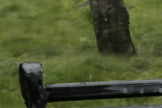} \\
  \includegraphics[width=0.16\linewidth]{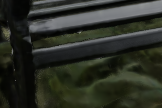} 
  \end{tabular}
}
&

\makecell[c]{\includegraphics[width=0.33\linewidth]{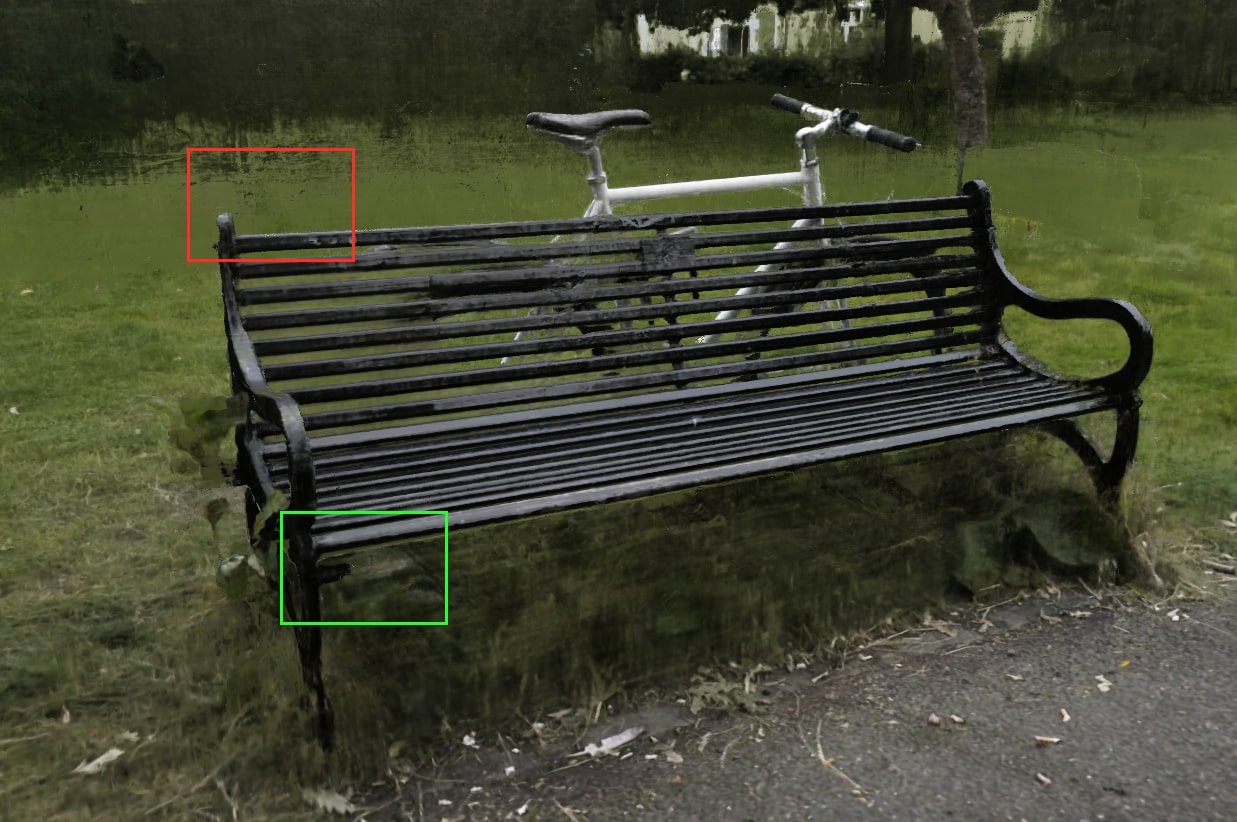}}

&
  
\makecell[c]{
  \begin{tabular}[t]{@{\,}c}
  \includegraphics[width=0.16\linewidth]{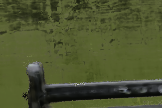} \\
  \includegraphics[width=0.16\linewidth]{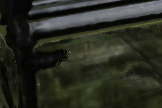} 
  \end{tabular}
}
\\
With low-resolution 3D grid & &
Without low-resolution 3D grid &


\end{tabular}
}
\caption{Visual comparison between MERF with and without a low-resolution 3D voxel grid (both models use three high-resolution 2D grids). Omitting the 3D grid often results in parts of the scene being poorly reconstructed; note the missing background in this example.}
\label{fig:visual_comparison_grid}
\end{figure*}

We begin our analysis with a comparison of MERF with SNeRG++: both use the same training pipeline, view-dependent appearance model, and rendering engine implementation, and this enables controlled experiments. To support our claim that MERF provides a favorable trade-off between memory consumption and rendering quality, we train models with varying resolutions: For MERF, we vary the feature plane resolution $R$ from $512$ and $3072$ while setting the grid resolution $L$ to either $512$ or $1024$. Likewise, for SNeRG++, we vary the grid resolution $L$ from $512$ to $2048$. To demonstrate the benefit of adding a low-resolution sparse 3D grid, we also train our models without the 3D grid. As can be seen in Figure~\ref{fig:vram_vs_psnr}, the memory consumption of SNeRG++ quickly rises to multiple gigabytes, whereas our model scales better to higher resolutions. For models without 3D grids we observe that quality saturates well below the other models. In Figure~\ref{fig:visual_comparison_snerg} we see that our model ($R = 2048$, $L = 512$) achieves similar quality to SNeRG++, while requiring a fraction of the memory.
Additionally, in Figure~\ref{fig:visual_comparison_grid} we see that ablating the 3D grid from MERF leads to a significant loss in quality.

\subsection{Real-time Rendering Evaluation}
\begin{figure*}[t]
\centering
\resizebox{0.99\linewidth}{!}{
\begin{tabular}[!t]{@{}c@{}c@{}}

\makecell[c]{
  \includegraphics[width=0.2\linewidth]{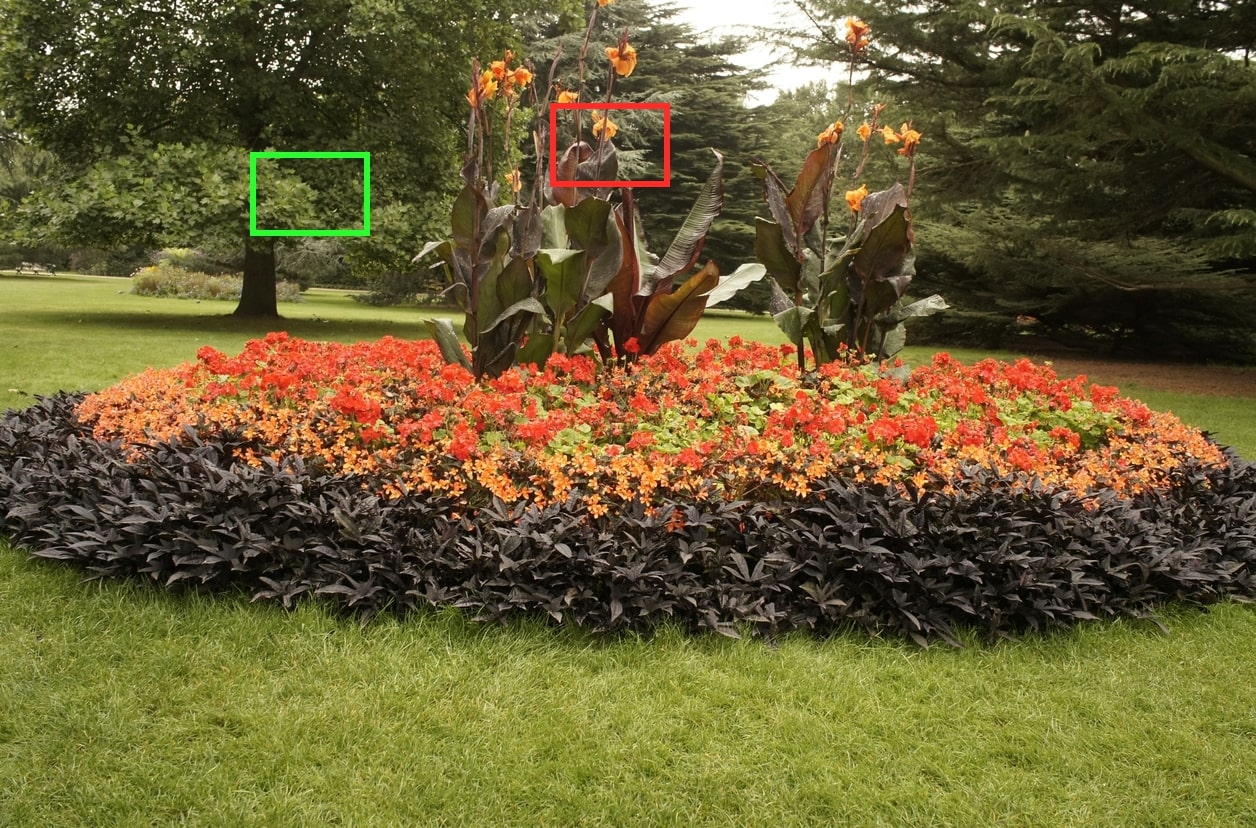}\\
  \texttt{flowerbed}
  } & 
  \begin{tabular}[t]{@{\,}c@{\,}c@{\,}c@{\,}c@{\,}c}
  \includegraphics[width=0.15\linewidth]{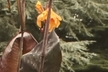} & 
  \includegraphics[width=0.15\linewidth]{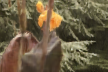} & 
  \includegraphics[width=0.15\linewidth]{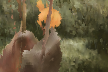} & 
  \includegraphics[width=0.15\linewidth]{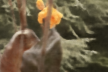} & 
  \includegraphics[width=0.15\linewidth]{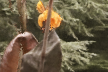} \\
  \includegraphics[width=0.15\linewidth]{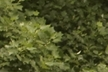} & 
  \includegraphics[width=0.15\linewidth]{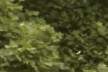} & 
  \includegraphics[width=0.15\linewidth]{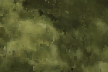} & 
  \includegraphics[width=0.15\linewidth]{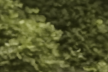} & 
  \includegraphics[width=0.15\linewidth]
  {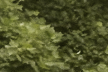} \\
  Ground truth & Ours & Mobile-NeRF & Instant NGP & mip-NeRF 360
  \end{tabular}
 \vspace{.5em} \\

\makecell[c]{
  \includegraphics[width=0.2\linewidth]{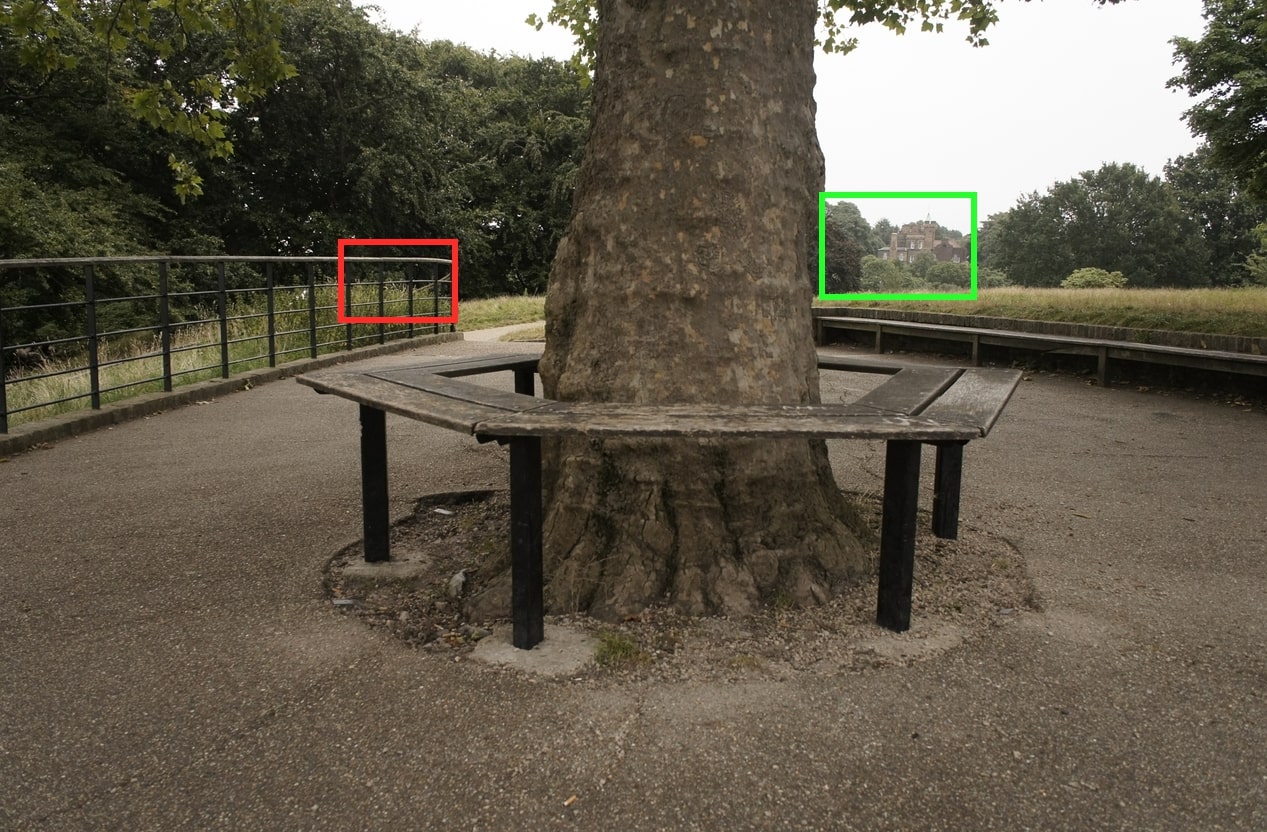}\\
  \texttt{treehill}
  } & 
  \begin{tabular}[t]{@{\,}c@{\,}c@{\,}c@{\,}c@{\,}c}
  \includegraphics[width=0.15\linewidth]{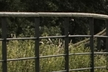} & 
  \includegraphics[width=0.15\linewidth]{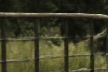} & 
  \includegraphics[width=0.15\linewidth]{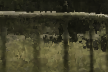} & 
  \includegraphics[width=0.15\linewidth]{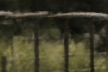} & 
  \includegraphics[width=0.15\linewidth]{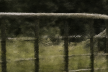} \\
  \includegraphics[width=0.15\linewidth]{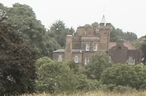} & 
  \includegraphics[width=0.15\linewidth]{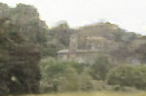} & 
  \includegraphics[width=0.15\linewidth]{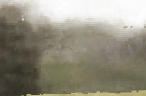} & 
  \includegraphics[width=0.15\linewidth]{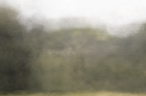} & 
  \includegraphics[width=0.15\linewidth]
  {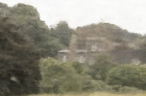} \\
  Ground truth & Ours & Mobile-NeRF & Instant NGP & mip-NeRF 360
  \end{tabular} \\

  
\end{tabular}
}
\caption{Visual comparison between MERF and other view synthesis methods. Mobile-NeRF~\cite{chen2022mobilenerf} is the only other real-time method (30fps or better). Instant NGP~\cite{muller2022instant} runs at interactive rates (around 5fps) and mip-NeRF 360~\cite{barron2022mipnerf360} is an extremely heavyweight offline method (around 30 seconds to render a single frame), representing the current state-of-the-art view synthesis quality.}
\label{fig:visual_comparison}
\end{figure*}

Finally, we evaluate the rendering speed of MERF, MobileNeRF, SNeRG++, and Instant-NGP in frames per second (FPS). Note that MERF, Mobile-NeRF and SNeRG++ all run in the browser and use the view-dependence model introduced by Hedman et al. \cite{hedman2021snerg}. In contrast, Instant-NGP uses a different view-dependence model, and is implemented in CUDA and is therefore less portable across devices. For benchmarking the methods that include web viewers (MERF, Mobile-NERF, SNeRG++) we use an M1 MacBook Pro and set the rendering resolution to $1280\times720$. When evaluating against Instant-NGP, to make the comparison fair we use an RTX 3090 GPU (which Instant-NGP is optimized for) and increase the rendering resolution to $1920\times1080$ to demonstrate MERF's scalability on high-powered devices. As can be seen in Table~\ref{tab:macbook}, our method runs faster than SNeRG++ while consuming only one fifth of the memory. While MobileNeRF achieves higher frame rates on the Macbook than MERF, it requires twice as much video memory and reduces rendering quality (a $1.24$ dB reduction in PSNR). This reduced quality is especially evident in background regions, as shown in Figure~\ref{fig:visual_comparison}. From our experiment with the RTX 3090, we see that Instant-NGP does not achieve real-time performance (4 FPS), while MERF renders at frame rates well above 100.

\subsection{Limitations}
Since we use the view-dependence model introduced in SNeRG~\cite{hedman2021snerg}, we also inherit its limitations: By evaluating view-dependent color once per ray, we are unable to faithfully model view-dependent appearance for rays that intersect with semi-transparent objects. Furthermore, since the tiny MLP has limited capacity, it may struggle to scale to much larger scenes or objects with complex reflections. 

Moreover, our method still performs volume rendering, which limits it to devices equipped with a sufficiently powerful GPU such as laptops, tablets or workstations. Running our model on smaller, thermally limited devices such as mobile phones or headsets will require further reductions in memory and runtime.

\section{Conclusion}

We have presented MERF, a compressed volume representation for radiance fields, which admits real-time rendering of large-scale scenes in a browser. By using novel hybrid volumetric parameterization, a novel contraction function that preserves straight lines, and a baking procedure that ensures that our real-time representation describes the same radiance field as was used during optimization, MERF is able to achieve faster and more accurate real-time rendering of large and complicated real-world scenes than prior real-time NeRF-like models. Out of all real-time methods, ours produces the highest-quality renderings for any given memory budget. Not only does it achieve $31.6\%$ (MSE) higher quality in the outdoor scenes compared to MobileNeRF, the previous state-of-the-art, it also requires less than half of the GPU memory.

\begin{acks}
We thank Marcos Seefelder, Julien Philip and Simon Rodriguez for their suggestions on shader optimization. This work was supported by the ERC Starting Grant LEGO3D (850533) and the DFG EXC number 2064/1 - project number 390727645.
\end{acks}

\begin{table}[t]
\centering
\begin{tabular}{@{}l|ccccc@{}}
& PSNR $\uparrow$ & SSIM $\uparrow$ & LPIPS $\downarrow$ \\ \hline
(a) Pre-baking &    23.20 &   0.620 &    0.336 \\
(b) w/o quant.-aware training & 22.64 & 0.603 &  0.347 \\
(c) Spherical contraction &  23.22 &    0.619 &    0.341 \\
Ours (Post-baking)  & 23.19 & 0.616 & 0.343 \\
\end{tabular}
\caption{We compare our final model after baking to the model before baking (a) demonstrating that our Proposal-MLP-aware baking pipeline is almost lossless. Omitting quantization-aware training (b) leads to a drop in rendering quality. Our proposed contraction function performs on par with the original spherical contraction function (c), while enabling efficient ray-AABB intersection tests. Results are averaged over all outdoor scenes.}
\label{tab:model_analysis}
\end{table}

\definecolor{RowColorA}{rgb}{1.,0.94,0.94}
\definecolor{RowColorB}{rgb}{0.94,0.94,1}

\begin{table}[t]
\centering
\resizebox{\linewidth}{!}{
\large
\begin{tabular}{l|cccccccc@{}}
& PSNR $\uparrow$ & SSIM $\uparrow$ & LPIPS $\downarrow$ & VRAM $\downarrow$ & DISK $\downarrow$ & FPS $\uparrow$\\ \hline
\rowcolor{RowColorA} \multicolumn{1}{c}{} & \multicolumn{6}{c}{MacBook M1 Pro, 1280$\times$720} \\ 
\rowcolor{RowColorA} Mobile-NeRF & 21.95 & 0.470 & 0.470 & 1162 & 345 &    \textbf{65.7} \\
\rowcolor{RowColorA} SNeRG++                               &    \textbf{23.64} &    \textbf{0.672} &    \textbf{0.285} & 4571 & 3785 & 18.7 \\
\rowcolor{RowColorA} Ours                                  & 23.19 & 0.616 & 0.343 &    \textbf{524} &    \textbf{188} & 28.3 \\
\hline 
\rowcolor{RowColorB} \multicolumn{1}{c}{} & \multicolumn{6}{c}{NVIDIA RTX 3090, 1920$\times$1080} \\ 
\rowcolor{RowColorB} Instant-NGP                                  & 22.90 & 0.566 & 0.371 &   $-$ &  \textbf{107}  & 4 \\
\rowcolor{RowColorB} Ours                                  & \textbf{23.19} & \textbf{0.616} &  \textbf{0.343} &   \textbf{524} &  188 & \textbf{119} \\
\end{tabular}
}
\caption{Performance comparison on the ``outdoor'' scenes.}
\label{tab:macbook}
\end{table}

\bibliographystyle{ACM-Reference-Format}
\bibliography{doc}


\begin{thebibliography}{62}


\ifx \showCODEN    \undefined \def \showCODEN     #1{\unskip}     \fi
\ifx \showDOI      \undefined \def \showDOI       #1{#1}\fi
\ifx \showISBNx    \undefined \def \showISBNx     #1{\unskip}     \fi
\ifx \showISBNxiii \undefined \def \showISBNxiii  #1{\unskip}     \fi
\ifx \showISSN     \undefined \def \showISSN      #1{\unskip}     \fi
\ifx \showLCCN     \undefined \def \showLCCN      #1{\unskip}     \fi
\ifx \shownote     \undefined \def \shownote      #1{#1}          \fi
\ifx \showarticletitle \undefined \def \showarticletitle #1{#1}   \fi
\ifx \showURL      \undefined \def \showURL       {\relax}        \fi
\providecommand\bibfield[2]{#2}
\providecommand\bibinfo[2]{#2}
\providecommand\natexlab[1]{#1}
\providecommand\showeprint[2][]{arXiv:#2}

\bibitem[Aliaga et~al\mbox{.}(2002)]%
        {aliaga2002sea}
\bibfield{author}{\bibinfo{person}{D.G. Aliaga}, \bibinfo{person}{T.
  Funkhouser}, \bibinfo{person}{D. Yanovsky}, {and} \bibinfo{person}{I.
  Carlbom}.} \bibinfo{year}{2002}\natexlab{}.
\newblock \showarticletitle{Sea of images}.
\newblock \bibinfo{journal}{\emph{IEEE Visualization}} (\bibinfo{year}{2002}).
\newblock


\bibitem[Attal et~al\mbox{.}(2022)]%
        {attal2022learning}
\bibfield{author}{\bibinfo{person}{Benjamin Attal}, \bibinfo{person}{Jia-Bin
  Huang}, \bibinfo{person}{Michael Zollh{\"o}fer}, \bibinfo{person}{Johannes
  Kopf}, {and} \bibinfo{person}{Changil Kim}.} \bibinfo{year}{2022}\natexlab{}.
\newblock \showarticletitle{Learning Neural Light Fields with Ray-Space
  Embedding Networks}.
\newblock \bibinfo{journal}{\emph{CVPR}} (\bibinfo{year}{2022}).
\newblock


\bibitem[Attal et~al\mbox{.}(2020)]%
        {attal2020matryodska}
\bibfield{author}{\bibinfo{person}{Benjamin Attal}, \bibinfo{person}{Selena
  Ling}, \bibinfo{person}{Aaron Gokaslan}, \bibinfo{person}{Christian
  Richardt}, {and} \bibinfo{person}{James Tompkin}.}
  \bibinfo{year}{2020}\natexlab{}.
\newblock \showarticletitle{{MatryODShka}: Real-time {6DoF} Video View
  Synthesis using Multi-Sphere Images}.
\newblock \bibinfo{journal}{\emph{ECCV}} (\bibinfo{year}{2020}).
\newblock


\bibitem[Barron et~al\mbox{.}(2022)]%
        {barron2022mipnerf360}
\bibfield{author}{\bibinfo{person}{Jonathan~T. Barron}, \bibinfo{person}{Ben
  Mildenhall}, \bibinfo{person}{Dor Verbin}, \bibinfo{person}{Pratul~P.
  Srinivasan}, {and} \bibinfo{person}{Peter Hedman}.}
  \bibinfo{year}{2022}\natexlab{}.
\newblock \showarticletitle{{Mip-NeRF 360}: Unbounded Anti-Aliased Neural
  Radiance Fields}.
\newblock \bibinfo{journal}{\emph{CVPR}} (\bibinfo{year}{2022}).
\newblock


\bibitem[Bengio et~al\mbox{.}(2013)]%
        {bengio2013estimating}
\bibfield{author}{\bibinfo{person}{Yoshua Bengio}, \bibinfo{person}{Nicholas
  L{\'e}onard}, {and} \bibinfo{person}{Aaron Courville}.}
  \bibinfo{year}{2013}\natexlab{}.
\newblock \showarticletitle{Estimating or propagating gradients through
  stochastic neurons for conditional computation}.
\newblock \bibinfo{journal}{\emph{arXiv:1308.3432}} (\bibinfo{year}{2013}).
\newblock


\bibitem[Broxton et~al\mbox{.}(2020)]%
        {broxton20}
\bibfield{author}{\bibinfo{person}{Michael Broxton}, \bibinfo{person}{John
  Flynn}, \bibinfo{person}{Ryan Overbeck}, \bibinfo{person}{Daniel Erickson},
  \bibinfo{person}{Peter Hedman}, \bibinfo{person}{Matthew DuVall},
  \bibinfo{person}{Jason Dourgarian}, \bibinfo{person}{Jay Busch},
  \bibinfo{person}{Matt Whalen}, {and} \bibinfo{person}{Paul Debevec}.}
  \bibinfo{year}{2020}\natexlab{}.
\newblock \showarticletitle{Immersive Light Field Video with a Layered Mesh
  Representation}.
\newblock \bibinfo{journal}{\emph{ACM Transactions on Graphics}}
  (\bibinfo{year}{2020}).
\newblock


\bibitem[Buehler et~al\mbox{.}(2001)]%
        {buehler01ulr}
\bibfield{author}{\bibinfo{person}{Chris Buehler}, \bibinfo{person}{Michael
  Bosse}, \bibinfo{person}{Leonard McMillan}, \bibinfo{person}{Steven Gortler},
  {and} \bibinfo{person}{Michael Cohen}.} \bibinfo{year}{2001}\natexlab{}.
\newblock \showarticletitle{Unstructured Lumigraph Rendering}.
\newblock \bibinfo{journal}{\emph{SIGGRAPH}} (\bibinfo{year}{2001}).
\newblock


\bibitem[Chan et~al\mbox{.}(2022)]%
        {chan2021eg3d}
\bibfield{author}{\bibinfo{person}{Eric~R. Chan}, \bibinfo{person}{Connor~Z.
  Lin}, \bibinfo{person}{Matthew~A. Chan}, \bibinfo{person}{Koki Nagano},
  \bibinfo{person}{Boxiao Pan}, \bibinfo{person}{Shalini~De Mello},
  \bibinfo{person}{Orazio Gallo}, \bibinfo{person}{Leonidas Guibas},
  \bibinfo{person}{Jonathan Tremblay}, \bibinfo{person}{Sameh Khamis},
  \bibinfo{person}{Tero Karras}, {and} \bibinfo{person}{Gordon Wetzstein}.}
  \bibinfo{year}{2022}\natexlab{}.
\newblock \showarticletitle{Efficient Geometry-aware {3D} Generative
  Adversarial Networks}.
\newblock \bibinfo{journal}{\emph{CVPR}} (\bibinfo{year}{2022}).
\newblock


\bibitem[Chaurasia et~al\mbox{.}(2013)]%
        {chaurasia2013superpixel}
\bibfield{author}{\bibinfo{person}{Gaurav Chaurasia}, \bibinfo{person}{Sylvain
  Duchene}, \bibinfo{person}{Olga Sorkine-Hornung}, {and}
  \bibinfo{person}{George Drettakis}.} \bibinfo{year}{2013}\natexlab{}.
\newblock \showarticletitle{Depth Synthesis and Local Warps for Plausible
  Image-Based Navigation}.
\newblock \bibinfo{journal}{\emph{ACM Transactions on Graphics}}
  (\bibinfo{year}{2013}).
\newblock


\bibitem[Chen et~al\mbox{.}(2022b)]%
        {chen2022tensorf}
\bibfield{author}{\bibinfo{person}{Anpei Chen}, \bibinfo{person}{Zexiang Xu},
  \bibinfo{person}{Andreas Geiger}, \bibinfo{person}{Jingyi Yu}, {and}
  \bibinfo{person}{Hao Su}.} \bibinfo{year}{2022}\natexlab{b}.
\newblock \showarticletitle{{TensoRF}: Tensorial Radiance Fields}.
\newblock \bibinfo{journal}{\emph{ECCV}} (\bibinfo{year}{2022}).
\newblock


\bibitem[Chen et~al\mbox{.}(2022a)]%
        {chen2022mobilenerf}
\bibfield{author}{\bibinfo{person}{Zhiqin Chen}, \bibinfo{person}{Thomas
  Funkhouser}, \bibinfo{person}{Peter Hedman}, {and} \bibinfo{person}{Andrea
  Tagliasacchi}.} \bibinfo{year}{2022}\natexlab{a}.
\newblock \showarticletitle{{MobileNeRF}: Exploiting the polygon rasterization
  pipeline for efficient neural field rendering on mobile architectures}.
\newblock \bibinfo{journal}{\emph{arXiv:2208.00277}} (\bibinfo{year}{2022}).
\newblock


\bibitem[Debevec et~al\mbox{.}(1998)]%
        {debevec1998efficient}
\bibfield{author}{\bibinfo{person}{Paul Debevec}, \bibinfo{person}{Yizhou Yu},
  {and} \bibinfo{person}{George Borshukov}.} \bibinfo{year}{1998}\natexlab{}.
\newblock \showarticletitle{Efficient view-dependent image-based rendering with
  projective texture-mapping}.
\newblock \bibinfo{journal}{\emph{EGSR}} (\bibinfo{year}{1998}).
\newblock


\bibitem[Deng and Tartaglione(2023)]%
        {deng2023compressing}
\bibfield{author}{\bibinfo{person}{Chenxi~Lola Deng} {and}
  \bibinfo{person}{Enzo Tartaglione}.} \bibinfo{year}{2023}\natexlab{}.
\newblock \showarticletitle{Compressing Explicit Voxel Grid Representations:
  Fast NeRFs Become Also Small}.
\newblock \bibinfo{journal}{\emph{WACV}} (\bibinfo{year}{2023}).
\newblock


\bibitem[DeVries et~al\mbox{.}(2021)]%
        {devries2021unconstrained}
\bibfield{author}{\bibinfo{person}{Terrance DeVries},
  \bibinfo{person}{Miguel~Angel Bautista}, \bibinfo{person}{Nitish Srivastava},
  \bibinfo{person}{Graham~W. Taylor}, {and} \bibinfo{person}{Joshua~M.
  Susskind}.} \bibinfo{year}{2021}\natexlab{}.
\newblock \showarticletitle{Unconstrained Scene Generation with Locally
  Conditioned Radiance Fields}.
\newblock \bibinfo{journal}{\emph{ICCV}} (\bibinfo{year}{2021}).
\newblock


\bibitem[Flynn et~al\mbox{.}(2019)]%
        {flynn2019deepview}
\bibfield{author}{\bibinfo{person}{John Flynn}, \bibinfo{person}{Michael
  Broxton}, \bibinfo{person}{Paul Debevec}, \bibinfo{person}{Matthew DuVall},
  \bibinfo{person}{Graham Fyffe}, \bibinfo{person}{Ryan Overbeck},
  \bibinfo{person}{Noah Snavely}, {and} \bibinfo{person}{Richard Tucker}.}
  \bibinfo{year}{2019}\natexlab{}.
\newblock \showarticletitle{{DeepView}: View synthesis with learned gradient
  descent}.
\newblock \bibinfo{journal}{\emph{CVPR}} (\bibinfo{year}{2019}).
\newblock


\bibitem[Garbin et~al\mbox{.}(2021)]%
        {garbin2021fastnerf}
\bibfield{author}{\bibinfo{person}{Stephan~J. Garbin}, \bibinfo{person}{Marek
  Kowalski}, \bibinfo{person}{Matthew Johnson}, \bibinfo{person}{Jamie
  Shotton}, {and} \bibinfo{person}{Julien Valentin}.}
  \bibinfo{year}{2021}\natexlab{}.
\newblock \showarticletitle{{FastNeRF}: High-Fidelity Neural Rendering at
  200FPS}.
\newblock \bibinfo{journal}{\emph{ICCV}} (\bibinfo{year}{2021}).
\newblock


\bibitem[Gortler et~al\mbox{.}(1996)]%
        {gortler1996lumigraph}
\bibfield{author}{\bibinfo{person}{Steven~J. Gortler}, \bibinfo{person}{Radek
  Grzeszczuk}, \bibinfo{person}{Richard Szeliski}, {and}
  \bibinfo{person}{Michael~F. Cohen}.} \bibinfo{year}{1996}\natexlab{}.
\newblock \showarticletitle{The lumigraph}.
\newblock \bibinfo{journal}{\emph{SIGGRAPH}} (\bibinfo{year}{1996}).
\newblock


\bibitem[Hedman et~al\mbox{.}(2018)]%
        {hedman2018deep}
\bibfield{author}{\bibinfo{person}{Peter Hedman}, \bibinfo{person}{Julien
  Philip}, \bibinfo{person}{True Price}, \bibinfo{person}{Jan-Michael Frahm},
  \bibinfo{person}{George Drettakis}, {and} \bibinfo{person}{Gabriel Brostow}.}
  \bibinfo{year}{2018}\natexlab{}.
\newblock \showarticletitle{Deep blending for free-viewpoint image-based
  rendering}.
\newblock \bibinfo{journal}{\emph{SIGGRAPH Asia}} (\bibinfo{year}{2018}).
\newblock


\bibitem[Hedman et~al\mbox{.}(2021)]%
        {hedman2021snerg}
\bibfield{author}{\bibinfo{person}{Peter Hedman}, \bibinfo{person}{Pratul~P.
  Srinivasan}, \bibinfo{person}{Ben Mildenhall}, \bibinfo{person}{Jonathan~T.
  Barron}, {and} \bibinfo{person}{Paul Debevec}.}
  \bibinfo{year}{2021}\natexlab{}.
\newblock \showarticletitle{Baking Neural Radiance Fields for Real-Time View
  Synthesis}.
\newblock \bibinfo{journal}{\emph{ICCV}} (\bibinfo{year}{2021}).
\newblock


\bibitem[Karnewar et~al\mbox{.}(2022)]%
        {karnewar2022relu}
\bibfield{author}{\bibinfo{person}{Animesh Karnewar}, \bibinfo{person}{Tobias
  Ritschel}, \bibinfo{person}{Oliver Wang}, {and} \bibinfo{person}{Niloy
  Mitra}.} \bibinfo{year}{2022}\natexlab{}.
\newblock \showarticletitle{{ReLU} fields: The little non-linearity that
  could}.
\newblock \bibinfo{journal}{\emph{SIGGRAPH}} (\bibinfo{year}{2022}).
\newblock


\bibitem[Kingma and Ba(2015)]%
        {kingma2015adam}
\bibfield{author}{\bibinfo{person}{Diederik~P. Kingma} {and}
  \bibinfo{person}{Jimmy Ba}.} \bibinfo{year}{2015}\natexlab{}.
\newblock \showarticletitle{Adam: {A} Method for Stochastic Optimization}. In
  \bibinfo{booktitle}{\emph{ICLR}}.
\newblock


\bibitem[Kurz et~al\mbox{.}(2022)]%
        {kurz2022adanerf}
\bibfield{author}{\bibinfo{person}{Andreas Kurz}, \bibinfo{person}{Thomas
  Neff}, \bibinfo{person}{Zhaoyang Lv}, \bibinfo{person}{Michael
  Zollh\"{o}fer}, {and} \bibinfo{person}{Markus Steinberger}.}
  \bibinfo{year}{2022}\natexlab{}.
\newblock \showarticletitle{AdaNeRF: Adaptive Sampling for Real-time Rendering
  of Neural Radiance Fields}.
\newblock \bibinfo{journal}{\emph{ECCV}} (\bibinfo{year}{2022}).
\newblock


\bibitem[Levoy and Hanrahan(1996)]%
        {levoy1996light}
\bibfield{author}{\bibinfo{person}{Marc Levoy} {and} \bibinfo{person}{Pat
  Hanrahan}.} \bibinfo{year}{1996}\natexlab{}.
\newblock \showarticletitle{Light field rendering}.
\newblock \bibinfo{journal}{\emph{SIGGRAPH}} (\bibinfo{year}{1996}).
\newblock


\bibitem[Li et~al\mbox{.}(2022b)]%
        {li2022compressing}
\bibfield{author}{\bibinfo{person}{Lingzhi Li}, \bibinfo{person}{Zhen Shen},
  \bibinfo{person}{Zhongshu Wang}, \bibinfo{person}{Li Shen}, {and}
  \bibinfo{person}{Liefeng Bo}.} \bibinfo{year}{2022}\natexlab{b}.
\newblock \showarticletitle{Compressing Volumetric Radiance Fields to 1 MB}.
\newblock \bibinfo{journal}{\emph{arXiv:2211.16386}} (\bibinfo{year}{2022}).
\newblock


\bibitem[Li et~al\mbox{.}(2022d)]%
        {li2022nerfacc}
\bibfield{author}{\bibinfo{person}{Ruilong Li}, \bibinfo{person}{Matthew
  Tancik}, {and} \bibinfo{person}{Angjoo Kanazawa}.}
  \bibinfo{year}{2022}\natexlab{d}.
\newblock \showarticletitle{NerfAcc: A General NeRF Accleration Toolbox.}
\newblock \bibinfo{journal}{\emph{arXiv:2210.04847}} (\bibinfo{year}{2022}).
\newblock


\bibitem[Li et~al\mbox{.}(2022a)]%
        {li2022steernerf}
\bibfield{author}{\bibinfo{person}{Sicheng Li}, \bibinfo{person}{Hao Li},
  \bibinfo{person}{Yue Wang}, \bibinfo{person}{Yiyi Liao}, {and}
  \bibinfo{person}{Lu Yu}.} \bibinfo{year}{2022}\natexlab{a}.
\newblock \showarticletitle{SteerNeRF: Accelerating NeRF Rendering via Smooth
  Viewpoint Trajectory}.
\newblock \bibinfo{journal}{\emph{arXiv:2212.08476}} (\bibinfo{year}{2022}).
\newblock


\bibitem[Li et~al\mbox{.}(2022c)]%
        {li2022neulf}
\bibfield{author}{\bibinfo{person}{Zhong Li}, \bibinfo{person}{Liangchen Song},
  \bibinfo{person}{Celong Liu}, \bibinfo{person}{Junsong Yuan}, {and}
  \bibinfo{person}{Yi Xu}.} \bibinfo{year}{2022}\natexlab{c}.
\newblock \showarticletitle{NeuLF: Efficient Novel View Synthesis with Neural
  4D Light Field}.
\newblock \bibinfo{journal}{\emph{EGSR}} (\bibinfo{year}{2022}).
\newblock


\bibitem[Lin et~al\mbox{.}(2022)]%
        {lin2022neurmips}
\bibfield{author}{\bibinfo{person}{Zhi-Hao Lin}, \bibinfo{person}{Wei-Chiu Ma},
  \bibinfo{person}{Hao-Yu Hsu}, \bibinfo{person}{Yu-Chiang~Frank Wang}, {and}
  \bibinfo{person}{Shenlong Wang}.} \bibinfo{year}{2022}\natexlab{}.
\newblock \showarticletitle{NeurMiPs: Neural Mixture of Planar Experts for View
  Synthesis}.
\newblock \bibinfo{journal}{\emph{CVPR}} (\bibinfo{year}{2022}).
\newblock


\bibitem[Lindell et~al\mbox{.}(2021)]%
        {lindell2021autoint}
\bibfield{author}{\bibinfo{person}{David~B. Lindell},
  \bibinfo{person}{Julien~N.P. Martel}, {and} \bibinfo{person}{Gordon
  Wetzstein}.} \bibinfo{year}{2021}\natexlab{}.
\newblock \showarticletitle{AutoInt: Automatic Integration for Fast Neural
  Rendering}.
\newblock \bibinfo{journal}{\emph{CVPR}} (\bibinfo{year}{2021}).
\newblock


\bibitem[Martin-Brualla et~al\mbox{.}(2021)]%
        {martinbrualla2020nerfw}
\bibfield{author}{\bibinfo{person}{Ricardo Martin-Brualla},
  \bibinfo{person}{Noha Radwan}, \bibinfo{person}{Mehdi S.~M. Sajjadi},
  \bibinfo{person}{Jonathan~T. Barron}, \bibinfo{person}{Alexey Dosovitskiy},
  {and} \bibinfo{person}{Daniel Duckworth}.} \bibinfo{year}{2021}\natexlab{}.
\newblock \showarticletitle{{NeRF} in the Wild: Neural Radiance Fields for
  Unconstrained Photo Collections}.
\newblock \bibinfo{journal}{\emph{CVPR}} (\bibinfo{year}{2021}).
\newblock


\bibitem[Max(1995)]%
        {max1995optical}
\bibfield{author}{\bibinfo{person}{Nelson Max}.}
  \bibinfo{year}{1995}\natexlab{}.
\newblock \showarticletitle{Optical models for direct volume rendering}.
\newblock \bibinfo{journal}{\emph{IEEE Transactions on Visualization and
  Computer Graphics}} (\bibinfo{year}{1995}).
\newblock


\bibitem[Mildenhall et~al\mbox{.}(2019)]%
        {mildenhall2019llff}
\bibfield{author}{\bibinfo{person}{Ben Mildenhall}, \bibinfo{person}{Pratul~P.
  Srinivasan}, \bibinfo{person}{Rodrigo Ortiz-Cayon},
  \bibinfo{person}{Nima~Khademi Kalantari}, \bibinfo{person}{Ravi Ramamoorthi},
  \bibinfo{person}{Ren Ng}, {and} \bibinfo{person}{Abhishek Kar}.}
  \bibinfo{year}{2019}\natexlab{}.
\newblock \showarticletitle{Local Light Field Fusion: Practical View Synthesis
  with Prescriptive Sampling Guidelines}.
\newblock \bibinfo{journal}{\emph{ACM Transactions on Graphics}}
  (\bibinfo{year}{2019}).
\newblock


\bibitem[Mildenhall et~al\mbox{.}(2020)]%
        {mildenhall2020nerf}
\bibfield{author}{\bibinfo{person}{Ben Mildenhall}, \bibinfo{person}{Pratul~P.
  Srinivasan}, \bibinfo{person}{Matthew Tancik}, \bibinfo{person}{Jonathan~T.
  Barron}, \bibinfo{person}{Ravi Ramamoorthi}, {and} \bibinfo{person}{Ren Ng}.}
  \bibinfo{year}{2020}\natexlab{}.
\newblock \showarticletitle{{NeRF}: Representing Scenes as Neural Radiance
  Fields for View Synthesis}.
\newblock \bibinfo{journal}{\emph{ECCV}} (\bibinfo{year}{2020}).
\newblock


\bibitem[M{\"u}ller et~al\mbox{.}(2022)]%
        {muller2022instant}
\bibfield{author}{\bibinfo{person}{Thomas M{\"u}ller}, \bibinfo{person}{Alex
  Evans}, \bibinfo{person}{Christoph Schied}, {and} \bibinfo{person}{Alexander
  Keller}.} \bibinfo{year}{2022}\natexlab{}.
\newblock \showarticletitle{Instant neural graphics primitives with a
  multiresolution hash encoding}.
\newblock \bibinfo{journal}{\emph{SIGGRAPH}} (\bibinfo{year}{2022}).
\newblock


\bibitem[Munkberg et~al\mbox{.}(2022)]%
        {munkberg2022extracting}
\bibfield{author}{\bibinfo{person}{Jacob Munkberg}, \bibinfo{person}{Jon
  Hasselgren}, \bibinfo{person}{Tianchang Shen}, \bibinfo{person}{Jun Gao},
  \bibinfo{person}{Wenzheng Chen}, \bibinfo{person}{Alex Evans},
  \bibinfo{person}{Thomas M{\"u}ller}, {and} \bibinfo{person}{Sanja Fidler}.}
  \bibinfo{year}{2022}\natexlab{}.
\newblock \showarticletitle{Extracting Triangular 3D Models, Materials, and
  Lighting From Images}.
\newblock \bibinfo{journal}{\emph{CVPR}} (\bibinfo{year}{2022}).
\newblock


\bibitem[Neff et~al\mbox{.}(2021)]%
        {neff2021donerf}
\bibfield{author}{\bibinfo{person}{Thomas Neff}, \bibinfo{person}{Pascal
  Stadlbauer}, \bibinfo{person}{Mathias Parger}, \bibinfo{person}{Andreas
  Kurz}, \bibinfo{person}{Joerg~H. Mueller}, \bibinfo{person}{Chakravarty
  R.~Alla Chaitanya}, \bibinfo{person}{Anton Kaplanyan}, {and}
  \bibinfo{person}{Markus Steinberger}.} \bibinfo{year}{2021}\natexlab{}.
\newblock \showarticletitle{{DONeRF}: Towards Real-Time Rendering of Compact
  Neural Radiance Fields using Depth Oracle Networks}.
\newblock \bibinfo{journal}{\emph{Computer Graphics Forum}}
  (\bibinfo{year}{2021}).
\newblock


\bibitem[Piala and Clark(2021)]%
        {piala2021terminerf}
\bibfield{author}{\bibinfo{person}{Martin Piala} {and} \bibinfo{person}{Ronald
  Clark}.} \bibinfo{year}{2021}\natexlab{}.
\newblock \showarticletitle{TermiNeRF: Ray Termination Prediction for Efficient
  Neural Rendering}.
\newblock \bibinfo{journal}{\emph{3DV}} (\bibinfo{year}{2021}).
\newblock


\bibitem[Reiser et~al\mbox{.}(2021)]%
        {reiser2021kilonerf}
\bibfield{author}{\bibinfo{person}{Christian Reiser}, \bibinfo{person}{Songyou
  Peng}, \bibinfo{person}{Yiyi Liao}, {and} \bibinfo{person}{Andreas Geiger}.}
  \bibinfo{year}{2021}\natexlab{}.
\newblock \showarticletitle{{KiloNeRF}: Speeding up neural radiance fields with
  thousands of tiny {MLPs}}.
\newblock \bibinfo{journal}{\emph{ICCV}} (\bibinfo{year}{2021}).
\newblock


\bibitem[Riegler and Koltun(2021)]%
        {riegler2021stable}
\bibfield{author}{\bibinfo{person}{Gernot Riegler} {and}
  \bibinfo{person}{Vladlen Koltun}.} \bibinfo{year}{2021}\natexlab{}.
\newblock \showarticletitle{Stable view synthesis}.
\newblock \bibinfo{journal}{\emph{CVPR}} (\bibinfo{year}{2021}).
\newblock


\bibitem[R{\"u}ckert et~al\mbox{.}(2022)]%
        {ruckert2021adop}
\bibfield{author}{\bibinfo{person}{Darius R{\"u}ckert}, \bibinfo{person}{Linus
  Franke}, {and} \bibinfo{person}{Marc Stamminger}.}
  \bibinfo{year}{2022}\natexlab{}.
\newblock \showarticletitle{{ADOP}: Approximate differentiable one-pixel point
  rendering}.
\newblock \bibinfo{journal}{\emph{SIGGRAPH}} (\bibinfo{year}{2022}).
\newblock


\bibitem[Sitzmann et~al\mbox{.}(2021)]%
        {sitzmann2021lfns}
\bibfield{author}{\bibinfo{person}{Vincent Sitzmann}, \bibinfo{person}{Semon
  Rezchikov}, \bibinfo{person}{William~T. Freeman}, \bibinfo{person}{Joshua~B.
  Tenenbaum}, {and} \bibinfo{person}{Fredo Durand}.}
  \bibinfo{year}{2021}\natexlab{}.
\newblock \showarticletitle{Light Field Networks: Neural Scene Representations
  with Single-Evaluation Rendering}.
\newblock \bibinfo{journal}{\emph{NeurIPS}} (\bibinfo{year}{2021}).
\newblock


\bibitem[Songyou~Peng(2020)]%
        {peng2020convolutional}
\bibfield{author}{\bibinfo{person}{Lars Mescheder Marc Pollefeys Andreas~Geiger
  Songyou~Peng, Michael~Niemeyer}.} \bibinfo{year}{2020}\natexlab{}.
\newblock \showarticletitle{Convolutional Occupancy Networks}.
\newblock \bibinfo{journal}{\emph{ECCV}} (\bibinfo{year}{2020}).
\newblock


\bibitem[Sun et~al\mbox{.}(2022)]%
        {sun2022direct}
\bibfield{author}{\bibinfo{person}{Cheng Sun}, \bibinfo{person}{Min Sun}, {and}
  \bibinfo{person}{Hwann-Tzong Chen}.} \bibinfo{year}{2022}\natexlab{}.
\newblock \showarticletitle{Direct voxel grid optimization: Super-fast
  convergence for radiance fields reconstruction}.
\newblock \bibinfo{journal}{\emph{CVPR}} (\bibinfo{year}{2022}).
\newblock


\bibitem[Takikawa et~al\mbox{.}(2022)]%
        {takikawa2022variable}
\bibfield{author}{\bibinfo{person}{Towaki Takikawa}, \bibinfo{person}{Alex
  Evans}, \bibinfo{person}{Jonathan Tremblay}, \bibinfo{person}{Thomas
  M\"{u}ller}, \bibinfo{person}{Morgan McGuire}, \bibinfo{person}{Alec
  Jacobson}, {and} \bibinfo{person}{Sanja Fidler}.}
  \bibinfo{year}{2022}\natexlab{}.
\newblock \showarticletitle{Variable Bitrate Neural Fields}.
\newblock \bibinfo{journal}{\emph{ACM Transactions on Graphics}}
  (\bibinfo{year}{2022}).
\newblock


\bibitem[Tancik et~al\mbox{.}(2022)]%
        {tancik2022blocknerf}
\bibfield{author}{\bibinfo{person}{Matthew Tancik}, \bibinfo{person}{Vincent
  Casser}, \bibinfo{person}{Xinchen Yan}, \bibinfo{person}{Sabeek Pradhan},
  \bibinfo{person}{Ben Mildenhall}, \bibinfo{person}{Pratul Srinivasan},
  \bibinfo{person}{Jonathan~T. Barron}, {and} \bibinfo{person}{Henrik
  Kretzschmar}.} \bibinfo{year}{2022}\natexlab{}.
\newblock \showarticletitle{{Block-NeRF}: Scalable Large Scene Neural View
  Synthesis}.
\newblock \bibinfo{journal}{\emph{CVPR}} (\bibinfo{year}{2022}).
\newblock


\bibitem[Tewari et~al\mbox{.}(2022)]%
        {tewari2022advances}
\bibfield{author}{\bibinfo{person}{Ayush Tewari}, \bibinfo{person}{Justus
  Thies}, \bibinfo{person}{Ben Mildenhall}, \bibinfo{person}{Pratul
  Srinivasan}, \bibinfo{person}{Edgar Tretschk}, \bibinfo{person}{W Yifan},
  \bibinfo{person}{Christoph Lassner}, \bibinfo{person}{Vincent Sitzmann},
  \bibinfo{person}{Ricardo Martin-Brualla}, \bibinfo{person}{Stephen Lombardi},
  {et~al\mbox{.}}} \bibinfo{year}{2022}\natexlab{}.
\newblock \showarticletitle{Advances in neural rendering}.
\newblock \bibinfo{journal}{\emph{Computer Graphics Forum}}
  (\bibinfo{year}{2022}).
\newblock


\bibitem[Turki et~al\mbox{.}(2022)]%
        {turki2022meganerf}
\bibfield{author}{\bibinfo{person}{Haithem Turki}, \bibinfo{person}{Deva
  Ramanan}, {and} \bibinfo{person}{Mahadev Satyanarayanan}.}
  \bibinfo{year}{2022}\natexlab{}.
\newblock \showarticletitle{Mega-NERF: Scalable Construction of Large-Scale
  NeRFs for Virtual Fly-Throughs}.
\newblock \bibinfo{journal}{\emph{CVPR}} (\bibinfo{year}{2022}).
\newblock


\bibitem[Wang et~al\mbox{.}(2022b)]%
        {wang2022r2l}
\bibfield{author}{\bibinfo{person}{Huan Wang}, \bibinfo{person}{Jian Ren},
  \bibinfo{person}{Zeng Huang}, \bibinfo{person}{Kyle Olszewski},
  \bibinfo{person}{Menglei Chai}, \bibinfo{person}{Yun Fu}, {and}
  \bibinfo{person}{Sergey Tulyakov}.} \bibinfo{year}{2022}\natexlab{b}.
\newblock \showarticletitle{R2L: Distilling Neural Radiance Field to Neural
  Light Field for Efficient Novel View Synthesis}.
\newblock \bibinfo{journal}{\emph{ECCV}} (\bibinfo{year}{2022}).
\newblock


\bibitem[Wang et~al\mbox{.}(2021)]%
        {wang2021neus}
\bibfield{author}{\bibinfo{person}{Peng Wang}, \bibinfo{person}{Lingjie Liu},
  \bibinfo{person}{Yuan Liu}, \bibinfo{person}{Christian Theobalt},
  \bibinfo{person}{Taku Komura}, {and} \bibinfo{person}{Wenping Wang}.}
  \bibinfo{year}{2021}\natexlab{}.
\newblock \showarticletitle{{NeuS}: Learning Neural Implicit Surfaces by Volume
  Rendering for Multi-view Reconstruction}.
\newblock \bibinfo{journal}{\emph{NeurIPS}} (\bibinfo{year}{2021}).
\newblock


\bibitem[Wang et~al\mbox{.}(2004)]%
        {wang2004image}
\bibfield{author}{\bibinfo{person}{Zhou Wang}, \bibinfo{person}{Alan~C. Bovik},
  \bibinfo{person}{Hamid~R. Sheikh}, {and} \bibinfo{person}{Eero~P.
  Simoncelli}.} \bibinfo{year}{2004}\natexlab{}.
\newblock \showarticletitle{Image quality assessment: from error visibility to
  structural similarity}.
\newblock \bibinfo{journal}{\emph{IEEE TIP}} (\bibinfo{year}{2004}).
\newblock


\bibitem[Wang et~al\mbox{.}(2022a)]%
        {wang20224knerf}
\bibfield{author}{\bibinfo{person}{Zhongshu Wang}, \bibinfo{person}{Lingzhi
  Li}, \bibinfo{person}{Zhen Shen}, \bibinfo{person}{Li Shen}, {and}
  \bibinfo{person}{Liefeng Bo}.} \bibinfo{year}{2022}\natexlab{a}.
\newblock \showarticletitle{4K-NeRF: High Fidelity Neural Radiance Fields at
  Ultra High Resolutions}.
\newblock \bibinfo{journal}{\emph{arXiv:2212.04701}} (\bibinfo{year}{2022}).
\newblock


\bibitem[Wizadwongsa et~al\mbox{.}(2021)]%
        {wizadwongsa2021nex}
\bibfield{author}{\bibinfo{person}{Suttisak Wizadwongsa},
  \bibinfo{person}{Pakkapon Phongthawee}, \bibinfo{person}{Jiraphon
  Yenphraphai}, {and} \bibinfo{person}{Supasorn Suwajanakorn}.}
  \bibinfo{year}{2021}\natexlab{}.
\newblock \showarticletitle{{NeX}: Real-time View Synthesis with Neural Basis
  Expansion}.
\newblock \bibinfo{journal}{\emph{CVPR}} (\bibinfo{year}{2021}).
\newblock


\bibitem[Wu et~al\mbox{.}(2022a)]%
        {wu2022diver}
\bibfield{author}{\bibinfo{person}{Liwen Wu}, \bibinfo{person}{Jae~Yong Lee},
  \bibinfo{person}{Anand Bhattad}, \bibinfo{person}{Yuxiong Wang}, {and}
  \bibinfo{person}{David Forsyth}.} \bibinfo{year}{2022}\natexlab{a}.
\newblock \showarticletitle{DIVeR: Real-time and Accurate Neural Radiance
  Fields with Deterministic Integration for Volume Rendering}.
\newblock \bibinfo{journal}{\emph{CVPR}} (\bibinfo{year}{2022}).
\newblock


\bibitem[Wu et~al\mbox{.}(2022b)]%
        {wu2022snisr}
\bibfield{author}{\bibinfo{person}{Xiuchao Wu}, \bibinfo{person}{Jiamin Xu},
  \bibinfo{person}{Zihan Zhu}, \bibinfo{person}{Hujun Bao},
  \bibinfo{person}{Qixing Huang}, \bibinfo{person}{James Tompkin}, {and}
  \bibinfo{person}{Weiwei Xu}.} \bibinfo{year}{2022}\natexlab{b}.
\newblock \showarticletitle{Scalable Neural Indoor Scene Rendering}.
\newblock \bibinfo{journal}{\emph{ACM TOG}} (\bibinfo{year}{2022}).
\newblock


\bibitem[Yariv et~al\mbox{.}(2021)]%
        {yariv2021volume}
\bibfield{author}{\bibinfo{person}{Lior Yariv}, \bibinfo{person}{Jiatao Gu},
  \bibinfo{person}{Yoni Kasten}, {and} \bibinfo{person}{Yaron Lipman}.}
  \bibinfo{year}{2021}\natexlab{}.
\newblock \showarticletitle{Volume rendering of neural implicit surfaces}.
\newblock \bibinfo{journal}{\emph{NeurIPS}} (\bibinfo{year}{2021}).
\newblock


\bibitem[Yin et~al\mbox{.}(2019)]%
        {yin2019understanding}
\bibfield{author}{\bibinfo{person}{Penghang Yin}, \bibinfo{person}{Jiancheng
  Lyu}, \bibinfo{person}{Shuai Zhang}, \bibinfo{person}{Stanley Osher},
  \bibinfo{person}{Yingyong Qi}, {and} \bibinfo{person}{Jack Xin}.}
  \bibinfo{year}{2019}\natexlab{}.
\newblock \showarticletitle{Understanding straight-through estimator in
  training activation quantized neural nets}.
\newblock \bibinfo{journal}{\emph{ICLR}} (\bibinfo{year}{2019}).
\newblock


\bibitem[Yu et~al\mbox{.}(2022)]%
        {yu2021plenoxels}
\bibfield{author}{\bibinfo{person}{Alex Yu}, \bibinfo{person}{Sara
  Fridovich-Keil}, \bibinfo{person}{Matthew Tancik}, \bibinfo{person}{Qinhong
  Chen}, \bibinfo{person}{Benjamin Recht}, {and} \bibinfo{person}{Angjoo
  Kanazawa}.} \bibinfo{year}{2022}\natexlab{}.
\newblock \showarticletitle{Plenoxels: Radiance fields without neural
  networks}.
\newblock \bibinfo{journal}{\emph{CVPR}} (\bibinfo{year}{2022}).
\newblock


\bibitem[Yu et~al\mbox{.}(2021)]%
        {yu2021plenoctrees}
\bibfield{author}{\bibinfo{person}{Alex Yu}, \bibinfo{person}{Ruilong Li},
  \bibinfo{person}{Matthew Tancik}, \bibinfo{person}{Hao Li},
  \bibinfo{person}{Ren Ng}, {and} \bibinfo{person}{Angjoo Kanazawa}.}
  \bibinfo{year}{2021}\natexlab{}.
\newblock \showarticletitle{{PlenOctrees} for real-time rendering of neural
  radiance fields}.
\newblock \bibinfo{journal}{\emph{ICCV}} (\bibinfo{year}{2021}).
\newblock


\bibitem[Zhang et~al\mbox{.}(2022)]%
        {zhang2022digging}
\bibfield{author}{\bibinfo{person}{Jian Zhang}, \bibinfo{person}{Jinchi Huang},
  \bibinfo{person}{Bowen Cai}, \bibinfo{person}{Huan Fu},
  \bibinfo{person}{Mingming Gong}, \bibinfo{person}{Chaohui Wang},
  \bibinfo{person}{Jiaming Wang}, \bibinfo{person}{Hongchen Luo},
  \bibinfo{person}{Rongfei Jia}, \bibinfo{person}{Binqiang Zhao}, {and}
  \bibinfo{person}{Xing Tang}.} \bibinfo{year}{2022}\natexlab{}.
\newblock \showarticletitle{Digging into Radiance Grid for Real-Time View
  Synthesis with Detail Preservation}.
\newblock \bibinfo{journal}{\emph{ECCV}} (\bibinfo{year}{2022}).
\newblock


\bibitem[Zhang et~al\mbox{.}(2020)]%
        {kaizhang2020}
\bibfield{author}{\bibinfo{person}{Kai Zhang}, \bibinfo{person}{Gernot
  Riegler}, \bibinfo{person}{Noah Snavely}, {and} \bibinfo{person}{Vladlen
  Koltun}.} \bibinfo{year}{2020}\natexlab{}.
\newblock \showarticletitle{{NeRF++}: Analyzing and Improving Neural Radiance
  Fields}.
\newblock \bibinfo{journal}{\emph{arXiv:2010.07492}} (\bibinfo{year}{2020}).
\newblock


\bibitem[Zhang et~al\mbox{.}(2018)]%
        {zhang2018lpips}
\bibfield{author}{\bibinfo{person}{Richard Zhang}, \bibinfo{person}{Phillip
  Isola}, \bibinfo{person}{Alexei~A. Efros}, \bibinfo{person}{Eli Shechtman},
  {and} \bibinfo{person}{Oliver Wang}.} \bibinfo{year}{2018}\natexlab{}.
\newblock \showarticletitle{The Unreasonable Effectiveness of Deep Features as
  a Perceptual Metric}.
\newblock \bibinfo{journal}{\emph{CVPR}} (\bibinfo{year}{2018}).
\newblock


\bibitem[Zhou et~al\mbox{.}(2018)]%
        {zhou2018stereomag}
\bibfield{author}{\bibinfo{person}{Tinghui Zhou}, \bibinfo{person}{Richard
  Tucker}, \bibinfo{person}{John Flynn}, \bibinfo{person}{Graham Fyffe}, {and}
  \bibinfo{person}{Noah Snavely}.} \bibinfo{year}{2018}\natexlab{}.
\newblock \showarticletitle{Stereo Magnification: Learning View Synthesis using
  Multiplane Images}.
\newblock \bibinfo{journal}{\emph{SIGGRAPH}} (\bibinfo{year}{2018}).
\newblock


\end{thebibliography}

\appendix
\noindent

\section{Training details}
All SNeRG++ and MERF models use the same training hyperparameters and architectures. We train for 25000 iterations with a batch size of $2^{16}$ pixels. A training batch is created by sampling pixels from all training images. We use the Adam \citep{kingma2015adam} optimizer with an exponentially decaying learning rate. The learning rate is warmed up during the first 100 iterations where it is increased from \num{1e-4} to \num{1e-2}. Then the learning rate is decayed to \num{1e-3}. Adam's hyperparameters $\beta_1$, $\beta_2$ and $\epsilon$ are set to \num{0.9}, \num{0.99} and \num{1e-15}, respectively. To regularize the hash grids we use a weight decay of \num{0.1} on its grid values.

\section{Architecture}
For parameterizing all grids (i.e. 3D voxel grids and 2D planes) we use an MLP with a multi-resolution hash encoding \citep{muller2022instant}. The hash encoding uses 20 levels and the individual hash tables have $2^{21}$ entries. Following \citet{muller2022instant}, hash collisions are resolved with a 2-layer MLP with 64 hidden units. This MLP outputs an 8-dimensional vector representing density, diffuse RGB and the 4-dimensional view-dependency feature vector. For our deferred view-dependency model we closely follow SNeRG \citep{hedman2021snerg} and use a 3-layer MLP with 16 hidden units. As in SNeRG viewing directions are encoded with 4 frequencies. Following MipNeRF360 \citep{barron2022mipnerf360} we use hierarchical sampling with three levels and therefore require two Proposal-MLPs. The Proposal-MLPs consist of 2 layers with 64 hidden units and use a hash encoding. Since a Proposoal-MLP merely needs to model coarse geometry, we use for the Proposal-MLPs' hash encodings only 10 levels, a maximum grid resolution of $512$ and a hash table size of $2^{16}$.

\section{Benchmarking}
For each test scene we define a camera pose that is used for benchmarking. Camera poses are set programmatically in each viewer. For fair comparison, we ensure that resolutions and camera intrinsics (i.e. field of view) are identical across viewers. We compute the average frame rate across 150 frames.

By default, Instant-NGP makes use of progressive upsampling, where the image is first rendered at a lower resolution. When the camera rests additional low resolutions images are rendered that are dynamically combined into the final high resolution image. We also implemented progressive rendering as part of our webviewer. Independent of the method progressive upsampling speeds up rendering in proportion to the ratio of target resolution and initial render resolution. For simplicity, we disable progressive rendering for benchmarking.

\end{document}